\documentclass[sn-mathphys-num]{sn-jnl}

\usepackage{graphicx}
\usepackage{multirow}
\usepackage{amsmath,amssymb,amsfonts}
\usepackage{amsthm}
\usepackage{mathrsfs}
\usepackage[title]{appendix}
\usepackage{xcolor}
\usepackage{textcomp}
\usepackage{manyfoot}
\usepackage{booktabs}
\usepackage{algorithm}
\usepackage{algorithmicx}
\usepackage{algpseudocode}
\usepackage{listings}

\newcommand{\tabincell}[2]{\begin{tabular}{@{}#1@{}}#2\end{tabular}}

\begin{document}

\title{An Online Automatic Modulation Classification Scheme Based on Isolation Distributional Kernel}

\author*[1]{\fnm{Xinpeng} \sur{Li}}\email{lixp@lamda.nju.edu.cn}
\equalcont{These authors contributed equally to this work.}

\author[1]{\fnm{Zile} \sur{Jiang}}\email{zilejian@usc.edu}
\equalcont{These authors contributed equally to this work.}

\author[1]{\fnm{Kai Ming} \sur{Ting}}\email{tingkm@nju.edu.cn}

\author[2]{\fnm{Ye} \sur{Zhu}}\email{ye.zhu@ieee.org}

\affil*[1]{\orgdiv{National Key Laboratory for Novel Software Technology}, \orgname{Nanjing University}, \orgaddress{\street{163 Xianlin Avenue}, \city{Nanjing}, \postcode{210023}, \state{Jiangsu}, \country{China}}}

\affil[2]{\orgdiv{School of Information Technology}, \orgname{Deakin University}, \orgaddress{\street{221 Burwood Highway}, \city{Melbourne}, \postcode{3125}, \state{Victoria}, \country{Australia}}}

\abstract{Automatic Modulation Classification (AMC), as a crucial technique in modern non-cooperative communication networks, plays a key role in various civil and military applications. However, existing AMC methods usually are complicated and can work in batch mode only due to their high computational complexity. This paper introduces a new online AMC scheme based on Isolation Distributional Kernel. Our method stands out in two aspects. Firstly, it is the first proposal to represent baseband signals using a distributional kernel. Secondly, it introduces a pioneering AMC technique that works well in online settings under realistic time-varying channel conditions. Through extensive experiments in online settings, we demonstrate the effectiveness of the proposed classifier. Our results indicate that the proposed approach outperforms existing baseline models, including two state-of-the-art deep learning classifiers. Moreover, it distinguishes itself as the first online classifier for AMC with linear time complexity, which marks a significant efficiency boost for real-time applications.}


\maketitle

\section{Introduction}

Automatic Modulation Classification (AMC) is a crucial technique in modern non-cooperative communication networks. An AMC algorithm enables a radio frequency receiver to automatically recognize the modulation format of received signals. AMC is first studied for its military applications in electronic warfare. For example, it helps to intercept signals between adversary units to decipher and recover enemies' communications \cite{zhu2015automatic,peng2023survey}.

With the rapid development of new generation (5G and beyond) communication technologies, AMC comes to play a significant role in civilian communication networks such as Cognitive Radio (CR) \cite{liang2011cognitive,nurelmadina2021systematic}. In a CR network, the limited spectrum resource is allocated flexibly and adaptively. The transmitter can change its modulation format to adapt to the current external channel condition to optimize transmission quality and bandwidth usage. The receiver needs to know which modulation format is currently used in order to decode the signals correctly. 
The information of the modulation format can be inserted into the transmitted signal frames. However, this approach occupies extra spectrum resources and therefore reduces the spectrum efficiency. 

This issue can be better resolved by adopting an AMC method. At the receiver side, AMC is used to continuously monitor the current modulation format, thus ensuring that the signals can be correctly demodulated and the transmitted data can be successfully conveyed. AMC eliminates the overhead of transmitting modulation format information in the network protocol, effectively increasing the overall capacity of a CR network.

Conventional AMC methods can be categorized into likelihood-based (LB) classifiers \cite{huan1995likelihood,hong2000bpsk,panagiotou2000likelihood} and feature-based (FB) classifiers \cite{hipp1986modulation,swami2000hierarchical,gardner1988cyclic,azzouz1995automatic,park2008automatic,hazza2013overview}. An LB classifier calculates the likelihood of each modulation hypothesis and  selects the one with the maximum likelihood as the classification result. The LB classifier has been proved to be optimal in terms of classification accuracy when perfect channel parameters are known to the receiver. However, the LB classifier has high computational complexity, and its performance is significantly degraded in real-world channel environments because the channel parameters cannot be precisely obtained. In contrast, FB classifiers decide the modulation format based on some engineered features of the received signals (e.g., spectral-based features, wavelet transformation-based features and high-order statistical features.). FB classifiers are usually easy to implement, but have lower accuracy compared with LB classifiers. 

Recently, deep learning (DL)-based classifiers have been proposed for AMC, e.g., \cite{wang2019data,teng2020accumulated,mao2021attentive,zhang2018automatic,zeng2019spectrum,li2019survey}. In general, DL classifiers are a form of FB classifiers because they extract features automatically via some means. A DL classifier trains an artificial neural network to perform the classification task. Different architectures of neural networks have been applied to AMC tasks. For example, one-dimensional and two-dimensional convolutional neural network (CNN) \cite{wang2019data,OShea2018over, teng2020accumulated,mao2021attentive} and recursive neural network (RNN) \cite{xu2020RNN}. DL classifiers have been shown to achieve promising classification performance even when the signals are very weak.

However, all of these DL classifiers have some common drawbacks which compromise their effectiveness and practicality in real-world applications in online settings:
\begin{itemize}
    \item The architecture of a DL classifier is extremely complicated. It is known to be time-consuming to tune the hyperparameters of a network.
    
    \item Existing DL models are batch learning algorithms which are trained on a fixed dataset and cannot perform an online model update to adapt to the changing channel environment. The time-varying channel conditions can distort the received signals in terms of amplitude, phase, etc. This can cause a mismatch of channel conditions between training data and testing data, making batch learning algorithms such as DL classifiers less effective for AMC. This issue is best handled by introducing an online model update mechanism. However, integrating such a mechanism would further increase the complexity of a DL classifier. For example, an online retraining mechanism is proposed to compensate for channel impairments \cite{teng2020accumulated}, but it requires extra network training and storage overhead. In summary, \textbf{the complexity of DL model is a key obstacle in designing a practical online AMC scheme}.

    \item Most of the signal representations currently employed in DL are \textbf{indirect representations of the signals} because some   conversion from the original signals is required in each of these representations.
\end{itemize}

\begin{figure}[!t]
\centering
\begin{minipage}[b]{\textwidth}
    \centering
    \includegraphics[width=\textwidth]{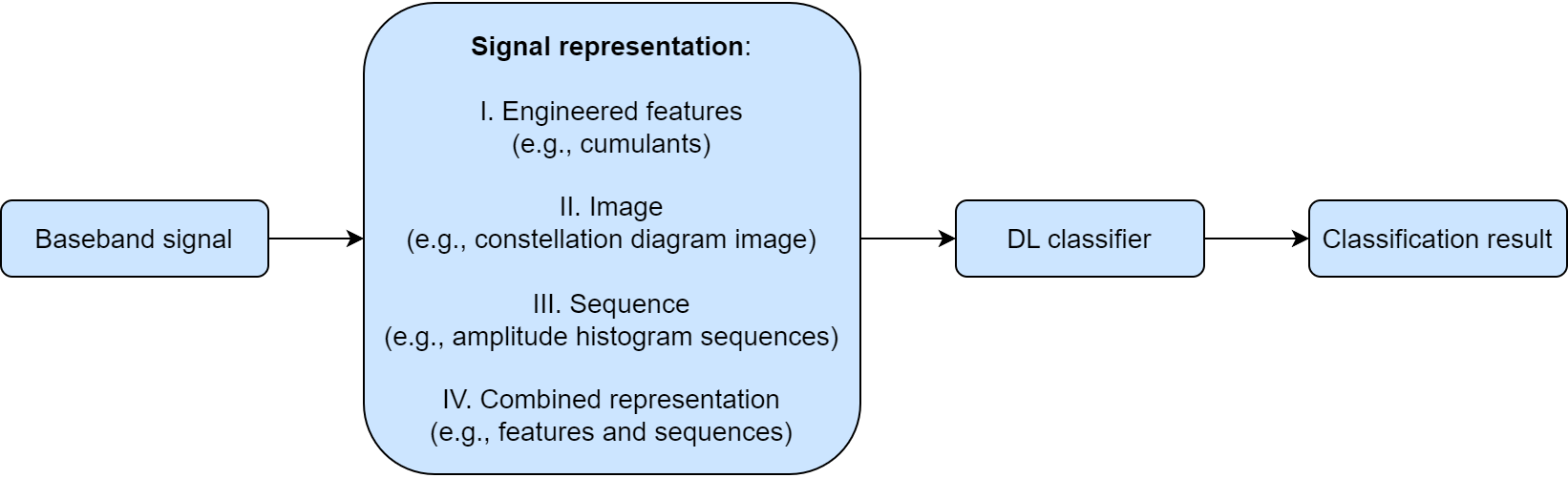}\par
    {\footnotesize\textbf{(a)} DL classifiers with four signal representations in batch mode}
\end{minipage}
\hfill
\vspace{-4pt}
\begin{minipage}[b]{\textwidth}
    \centering
    \includegraphics[width=\textwidth]{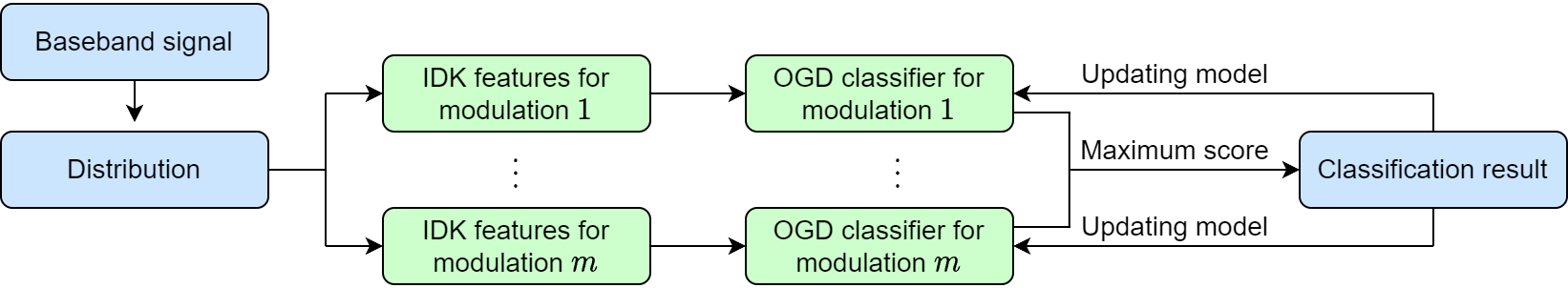}\par
    {\footnotesize\textbf{(b)} The proposed IDK-OGD classifier with distribution representation in online setting}
\end{minipage}
\caption{Functional diagrams of existing  DL classifiers and the proposed IDK-OGD classifier. The key differences are: (i) none of the existing signal representations used in DL classifiers include distribution representation; and (ii) DL classifiers are in batch mode and have difficulties in designing a practical online AMC scheme. The distribution representation, called Isolation Distribution Kernel (IDK), enables a kernel-based OGD classifier to be used readily for AMC}
\label{fig:diagram}
\end{figure}

To overcome above-mentioned issues, we propose a new online AMC scheme based on Isolation Distributional Kernel (IDK) \cite{ting2021-IDK} in this paper. Our contributions are:
\begin{enumerate}
    \item \textbf{An insight that the constellation diagram of a baseband signal sequence can be intuitively and effectively represented as a distribution, without loss of information.} This insight has enabled the use of a distributional measure to both represent baseband signals and perform similarity measurement between any two sets of baseband signals. This is an entirely new perspective in understanding the signals in AMC, apart from the four types of signal representations employed in existing DL classifiers, shown in Fig.~\hyperref[fig:diagram]{\ref{fig:diagram}(a)}.
    It is exactly because of the lack of this insight that no one has used a distribution measure in this way in AMC before.  
 
    \item With the above insight, we proposed to use a recent kernel called Isolation Distributional Kernel (IDK) \cite{ting2021-IDK} to represent each distribution exhibited in a constellation diagram. We show that IDK is a better representation than existing sequence and image representations as well as engineered features.
   
    \item Integrating IDK with Online Gradient Descent (OGD) \cite{OnlineKernelLearning-JMLR2016,Ting2021-X-Factor} to create an online AMC scheme, which is \textbf{the first online kernel classifier that works well in online AMC environments where there is a mismatch between the channel conditions of the training and testing data}. The proposed IDK-OGD classifier continuously updates its model based on new signal samples at a low time cost, which makes it especially efficient and effective in time-varying channel environments.
    
    \item Performing an extensive investigation comparing with state-of-the-art deep learning methods with different signal representations and other baselines. We conduct experiments on both synthetic datasets and an open source dataset RadioML2018.01A \cite{timothy2016RadioML}. In addition, various situations are taken into consideration including the mismatched channel conditions of training and testing data and time-varying channel conditions. These experiments \textbf{verify the effectiveness and practicality of the proposed AMC scheme}.
\end{enumerate}

The functional diagram of the proposed IDK-OGD classifier is shown in Fig.~\hyperref[fig:diagram]{\ref{fig:diagram}(b)}.

\begin{figure}[!t]
\centering
\includegraphics[width=0.7\textwidth]{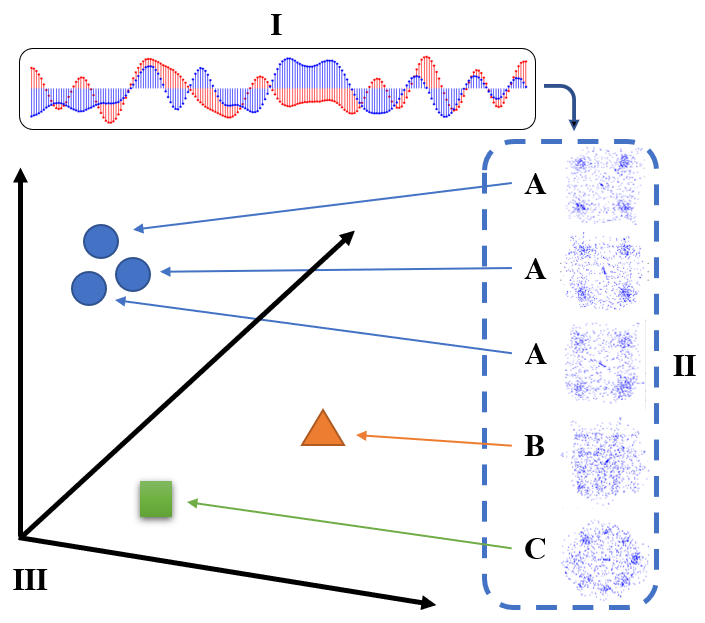}
\caption{Sketch of distribution-based signal representation (different letters, i.e., A, B \& C, stand for different modulation formats).  Each point-pair of the I/Q baseband signal (shown in \uppercase\expandafter{\romannumeral1}) is translated into a point in a two-dimensional I/Q signal space of a modulation format (shown in \uppercase\expandafter{\romannumeral2}). All the points in the I/Q space (as a sample set from an unknown distribution) can then be transformed via a distributional representation into a point in a feature space (shown in \uppercase\expandafter{\romannumeral3})}
\label{distribution}
\end{figure}

The distinctive characteristics of IDK-OGD can be discerned from each of its two key parts. First, the proposed use of IDK to represent signals in AMC is unique compared with existing representation methods in two aspects:
\begin{itemize}
    \item \textbf{Distribution-based signal representation}: 
    IDK perceives the in-phase (I) and quadrature (Q) components of a baseband signal, which are translated into a set of two-dimensional points shown in block II in Fig.~\ref{distribution}, as independent and identically distributed (i.i.d.) points sampled from an unknown distribution. Using these i.i.d. points, IDK automatically creates the features to represent this distribution. None of the conventional methods, including DL classifiers depicted in Fig.~\hyperref[fig:diagram]{\ref{fig:diagram}(a)}, have used distribution\footnote{Existing LB classifiers \cite{huan1995likelihood,hong2000bpsk,panagiotou2000likelihood} estimate distributions via a parametric method, and the parameters are determined \textit{a priori} and it is difficult to modify in accordance to the changing channel conditions. In addition, it is not a representation of signals. This is different from the non-parametric approach we proposed here. See Section~\ref{sec-distribution} for details.} as the representation of signals.

    \item  \textbf{No-learning method}: IDK is a no-learning method that employs no learned/engineered features, unlike deep learning and other existing methods which rely on learned/engineered features. IDK is also unique because it is the first application of the distributional kernel to AMC. 
\end{itemize}

Second, we show that the proposed IDK-OGD classifier is the first online AMC method that works well in time-varying channel conditions. This is due to the online model update capability of OGD which has been shown to work in the conventional data stream context \cite{OnlineKernelLearning-JMLR2016,Ting2021-X-Factor}. Most existing AMC methods can only work in batch mode, and even those which claimed to work in online mode are not readily applicable in an online setting (see the discussion section for a detailed exposition).

The rest of this paper is organized as follows. Section~\ref{sec-relatedwork} reviews the related work on DL-based AMC methods with different signal representations. The insight into using distribution as the representation of signals in AMC is provided in Section~\ref{sec-distribution}. We introduce the technical details of the proposed IDK-based online AMC scheme in Section~\ref{sec-IDK-OGD}. The experimental settings and results are presented in Section~\ref{sec-experiments}, and ablation studies are provided in Section~\ref{sec-ablationStudy}. Section~\ref{sec-runtime} presents a runtime comparison. A discussion of various issues is given in Section~\ref{sec-discussion}, and we conclude the paper in Section~\ref{sec-conclusion}.

\section{Related work}
\label{sec-relatedwork}

Over the past few years, with the remarkable advancement of deep learning (DL), numerous AMC methods based on DL have been proposed \cite{mao2018survey}. The DL classifiers have been categorized into four types based on their signal representations, i.e., engineered feature representation, image representation, sequence representation, and combined representation \cite{peng2023survey}. The functional diagram of the four categories is shown in Fig.~\hyperref[fig:diagram]{\ref{fig:diagram}(a)}. 
Here we focus on some related works in each of these four categories to examine how prior research has utilized different techniques to represent the original I/Q baseband signals, in order to contrast with our distribution representation.

\noindent
\textbf{Engineered feature representation}:
Cumulants are the most commonly used features which can be calculated from the original I/Q baseband signals, and they also find extensive application in DL-based AMC methods.
For example, Lee \textit{et al.} \cite{lee2019cumulants} introduced a DL classifier that can select effective higher-order cumulants for use in AMC, while discarding the ineffective ones.
This approach aims to enhance classification accuracy and simultaneously reduce computational complexity.
Spectral features constitute another prevalent feature type widely adopted in AMC, which can be derived from I/Q signals as well.
Lee \textit{et al.} \cite{lee2017spectral} proposed an AMC scheme that incorporates spectral features and cumulants to train a deep neural network (DNN) classifier.
Other engineered features used in DL-based AMC methods include circular features \cite{shah2019circular}, transform features \cite{wang2020wavelet} and cyclostationary features \cite{yin2023cyclostationary}.

However, these engineered features require manual extraction tailored to the candidate modulation formats and channel conditions, and they suffer from the loss of information during the conversion process from the original I/Q signals to the features.

\noindent
\textbf{Image representation}:
Motivated by the superior performance of DL in processing images, some researchers have attempted to convert the original I/Q baseband signals into images with a certain resolution to facilitate the application of DL in AMC.
Peng \textit{et al.} \cite{peng2019constellation} studied the use of constellation diagrams (i.e., scatter plots of I/Q signals in a two-dimensional I/Q space) as images in DL-based AMC methods, as well as the impact of image resolution and selected area size of the constellation diagram images on the classification accuracy.
Teng \textit{et al.} \cite{teng2020accumulated} transformed the original I/Q baseband signal into a scatter plot of polar coordinates and converted it into an image of $36 \times 36$ resolution for CNN classification.
Lee \textit{et al.} \cite{lee2019image} derived the features of I/Q signals such as cumulants, and then plotted these features on a two-dimensional complex plane to obtain feature images, which were then fed into a CNN classifier.
Other images such as eye diagram images \cite{wang2017eye}, amplitude spectrum of bispectrum (ASB) images \cite{li2019bispectrum} and ambiguity function images \cite{dai2016ambiguity} have also been used in DL-based AMC schemes.

Such AMC methods benefit from the powerful capability of DL in extracting features from images, but the image conversion process, from the original I/Q baseband signal, inevitably introduces noise and causes information loss. In short, the original signals are compromised during the conversion process.

\noindent
\textbf{Sequence representation} can be classified into two subcategories.
The first subcategory extracts features from baseband signal sequences in their original form (i.e., I/Q sequences).
The methods can be traced back to the work of O'Shea \textit{et al.} \cite{OShea2016IQ}, who introduced a joint training of feature extraction and one-dimensional CNN classifier on I/Q signal sequences.
Meng \textit{et al.} \cite{meng2018IQ} proposed to divide an I/Q signal sequence into several segments of successive subsequences to accommodate the varying lengths of I/Q sequences.
Furthermore, Dong \textit{et al.} \cite{dong2022IQ} designed a lightweight DNN classifier that can be deployed on edge devices for AMC, where the model uses a multichannel input to extract features from the I/Q signal sequences.

The second subcategory includes sequences derived from the original I/Q baseband signals through some processing, before the features are extracted.
For instance, Rajendran \textit{et al.} \cite{rajendran2018sequence} proposed to use Fast Fourier Transform (FFT) sequences, which can be derived by performing FFT on I/Q sequences, as the input of a long short-term memory (LSTM) network;
while Khan \textit{et al.} \cite{Khan2016histogram} obtained the amplitude histogram sequences of the original I/Q signals and used them to train a DL classifier.

The accuracy of the sequence representation methods hinges on how well the features can be extracted from sequences of the original I/Q signals or their derivatives. We show in the experiment section that a state-of-the-art  feature extraction method based on DL still has room for improvement.

\noindent
\textbf{Combined representation}:
The idea is to exploit the advantages of different signal representations.
Wang \textit{et al.} \cite{wang2019data} integrated I/Q sequences and constellation diagram images to tackle the problem of classifying high-order modulation formats.
Zhang \textit{et al.} \cite{zhang2019combined} proposed to utilize both smooth pseudo Wigner-Ville distribution (SPWVD) image and Born-Jordan distribution (BJD) image for CNN classification.

However, we contend that the advantages of individual representations are not well understood yet. Here we endeavor to better understand the strengths and weaknesses of the existing two core representations, i.e., image and sequence, with respect to the proposed distribution representation, which is described in the next section.

\section{Signals in distribution representation}
\label{sec-distribution}

Here we provide an insight of using distribution to represent signals, its advantages over image and sequence representations, and the details of the proposed representation in the following three subsections.

\subsection{The insight to represent a signal as a distribution in I/Q signal space}
\label{sec-insight}

In this article, we consider digital amplitude-phase (A/P) modulated signals, e.g., amplitude-shift keying (ASK), phase-shift keying (PSK), amplitude and phase-shift keying (APSK), and quadrature amplitude modulation (QAM). These modulation formats find widespread utilization in practical wireless communications and pose challenges for conventional AMC schemes.

\noindent
\textbf{Signals in I/Q signal space}: In general, a baseband signal $r(t)$ received from a wireless channel can be expressed as:
\begin{equation}
\label{eq:signal_model1}
    r(t) = s(t) * c(t) + w(t)
\end{equation}
\noindent
where $s(t)$ denotes the transmitted baseband signal, $c(t)$ is the channel impulse response, $w(t)$ represents the additive noise, and $*$ is a convolution operator.

The transmitted A/P modulated baseband signal $s(t)$ with $N$ symbols can be represented as:
\begin{equation}
\label{eq:signal_model2}
\begin{aligned}
    s(t) &= \sum_{n=1}^{N}a_n\exp(j\varphi_n)g(t-nT_s)
\end{aligned}
\end{equation}
and
\begin{equation}
\label{eq:signal_IQ}
\begin{aligned}
    a_n\exp(j\varphi_n) &= a_n\cos{\varphi_n}+j\cdot a_n\sin{\varphi_n}\\
    &=I_n+j\cdot Q_n
\end{aligned}
\end{equation}
\noindent
where $a_n$ and $\varphi_n$ are the amplitude and phase of the $n$-th symbol, respectively;  $g(t)$ signifies the pulse shaping filter; $T_s$ is the symbol duration; and $I_n$ and $Q_n$ denote the in-phase (I) and quadrature (Q) components of the $n$-th symbol, respectively.

In essence, the difference between two distinct A/P modulation formats lies in the arrangements of A/P pairs $(a_n, \varphi_n)$ of all $N$ symbols \cite{mao2021attentive}. 

As shown in equation (\ref{eq:signal_IQ}), an A/P modulated signal consists of a sequence of complex data points, of which the real and imaginary components are the in-phase (I) and quadrature (Q) components of the signal, respectively (as shown in block I in Fig.~\ref{distribution}).
This sequence can be represented as a set of points in a two-dimensional complex space of I and Q axes.
This set of points in the I/Q space is commonly referred to as a constellation diagram \cite{carvalho2013constellation}.
With equation (\ref{eq:signal_IQ}), it is an obvious fact that the constellation diagram of an A/P modulated signal encompasses all the amplitude and phase information of the signal, thus constellation diagrams are equivalent to the original signal sequences within the context of AMC, without loss of information.

Our insight is that \textbf{the points in the I/Q space (or constellation diagram) can be treated as i.i.d. points $x$ generated from an unknown distribution $\mathcal{P}$ in the I/Q space, i.e., $x \sim \mathcal{P}$}.

In other words, a signal $s(t)$ in I/Q components  is a collection $S$ of points in the I/Q space (denoted as signal $S$ hereafter); and a signal $S$ of one modulation format corresponds to an unknown distribution $\mathcal{P}_S$. 

Thus, the task of AMC in identifying the unknown modulation format of a signal $S$  can then be achieved by maximizing the distribution $\mathcal{P}_S$ with the distribution  $\mathcal{P}_{S_{\alpha}}$ of a modulation format $\alpha$ out of $m$ candidate modulation formats $\{M_1,M_2,\cdots,M_m\}$. It is formally expressed as:
\begin{equation}
\label{AMC}
 \text{AMC($S$)} = \underset{\alpha \in \{M_1,M_2,\cdots,M_m\}}{\arg\max} \mathcal{K}(\mathcal{P}_S, \mathcal{P}_{S_{\alpha}})
\end{equation}
\noindent
where $\mathcal{K}$ is a distributional kernel which measures the similarity between two distributions; and $S_\alpha$ is the i.i.d. sample set of a signal of modulation format $\alpha$.

Equation (\ref{AMC}) returns the modulation format $\alpha$ of $\mathcal{P}_{S_\alpha}$ which is most similar to $\mathcal{P}_S$.

This task is made simple by using a distributional representation with an injective mapping guarantee \cite{KernelMeanEmbedding2017, ting2021-IDK} such that I/Q baseband signals of the same modulation format are mapped into the same region in the feature space, while those of different modulation formats are mapped into different regions, as shown in Fig.~\ref{distribution} (from II to III). 

The above is a simplified conceptual illustration of the core idea. An actual implementation requires a classifier which employs $\mathcal{K}$ to measure the similarity between two distributions.  We employ a recent distribution kernel to perform the distribution-based representation/mapping, which is described in Section~\ref{sec-IDK}.

In the following subsection, we explain the reason why treating I/Q signals as distributions is a better representation than the existing image and sequence representations.

\subsection{Distribution representation versus image and sequence representations}

It is worth noting that the choice of signal representation plays a pivotal role in determining the performance of an AMC classifier. The sequence and image representations utilized in DL necessitate the transformation of the original signals into some forms of sequences or images as well as feature extraction. Either of these processes may lead to unnecessary information loss and the introduction of noise. 

In contrast, the distribution representation is a direct method that preserves the original signals within their native I/Q domains in a two-dimensional I/Q signal space, thus eliminating the need for conversion and its associated information loss and corruption. 

\begin{figure}[t]
\centering
\includegraphics[width=0.7\textwidth]{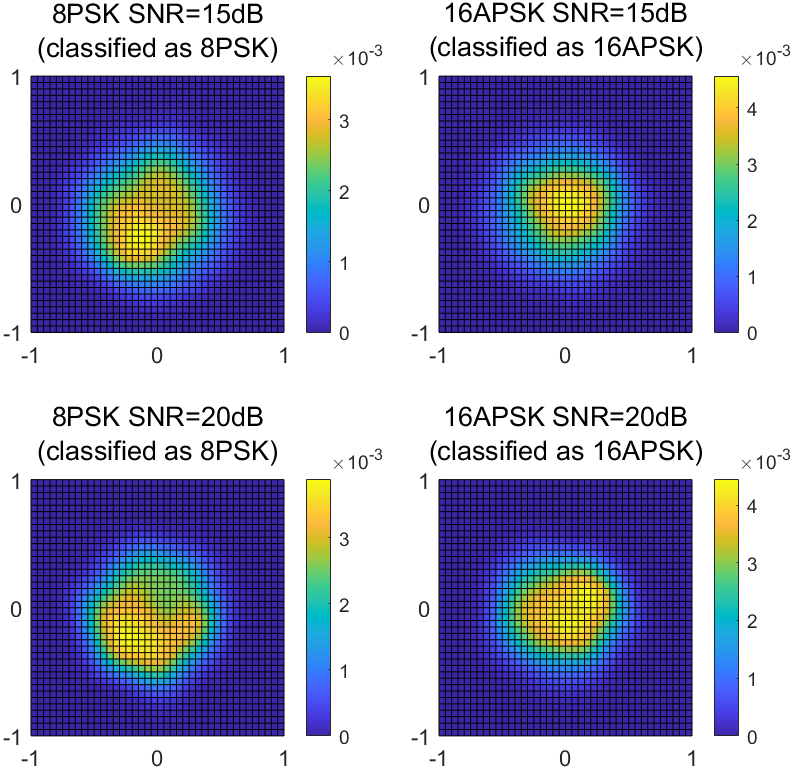}
\caption{Baseband signals of 8PSK \& 16APSK at SNR = 15 \& 20 dB in distribution representation}
\label{fig:robustness1}
\end{figure}

\begin{figure}[t]
\centering
\includegraphics[width=0.5\textwidth]{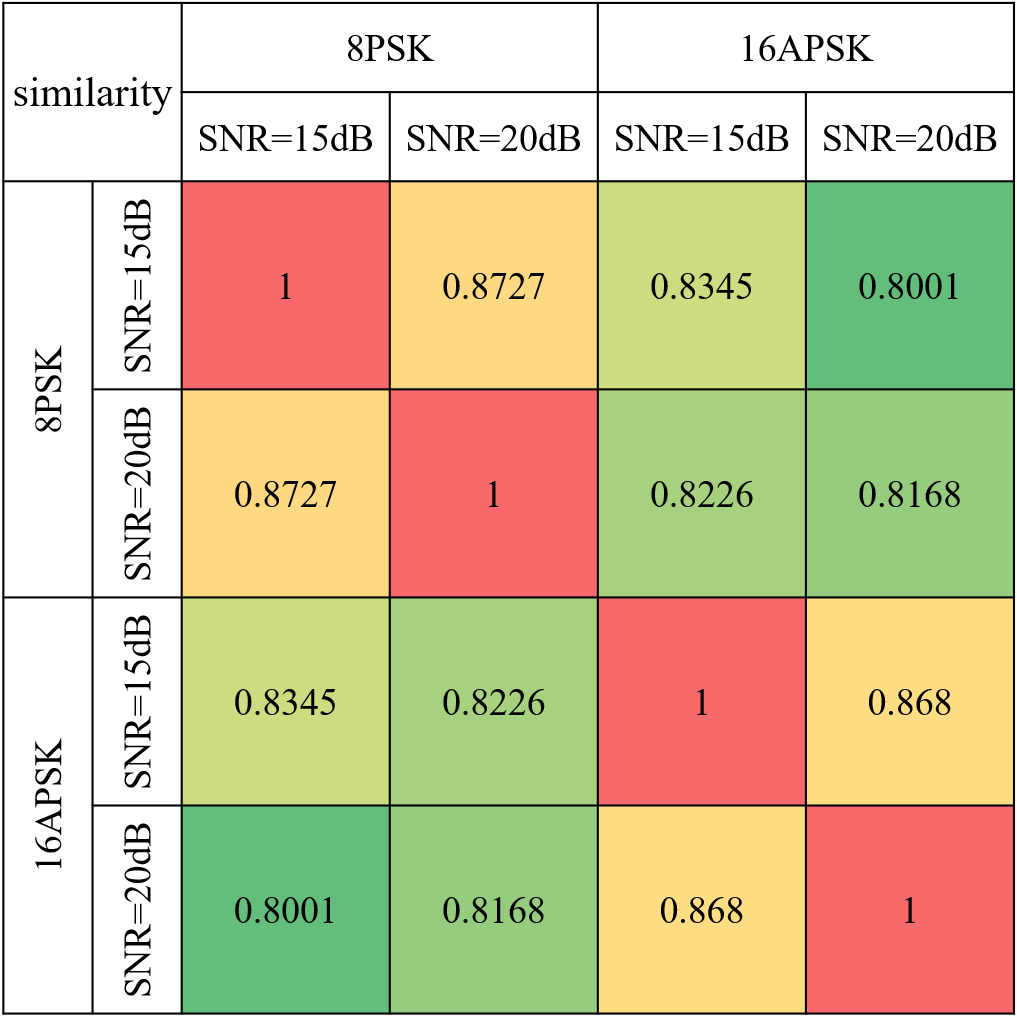}
\caption{The similarity matrix of the baseband signals, computed using a distributional kernel \cite{ting2021-IDK} when signals are in distribution representation}
\label{fig:similarity}
\end{figure}

\begin{figure}[t]
\centering
\includegraphics[width=0.7\textwidth]{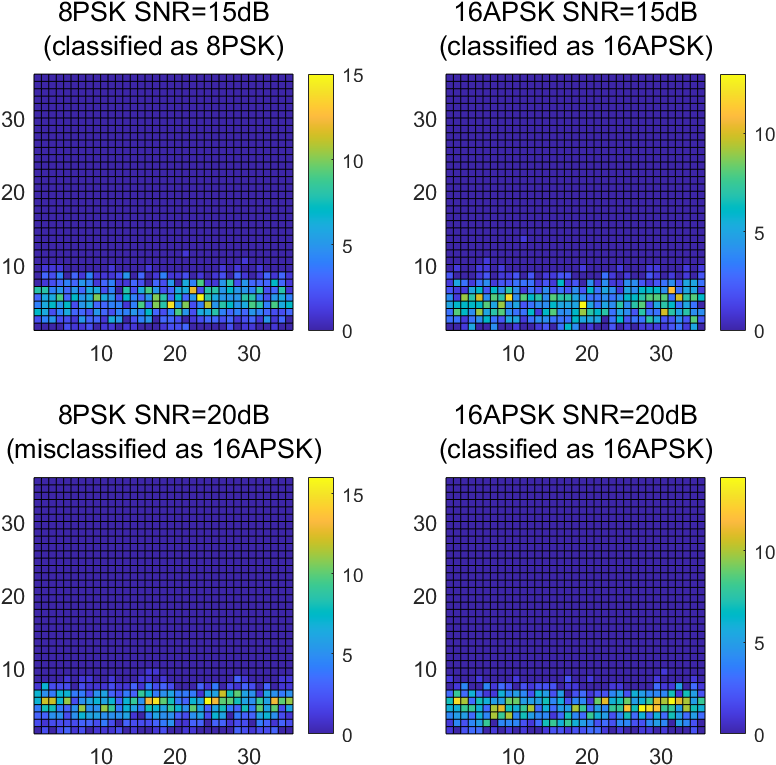}
\caption{Baseband signals in image representation}
\label{fig:robustness2}
\end{figure}

\begin{figure}[t]
\centering
\includegraphics[width=\textwidth]{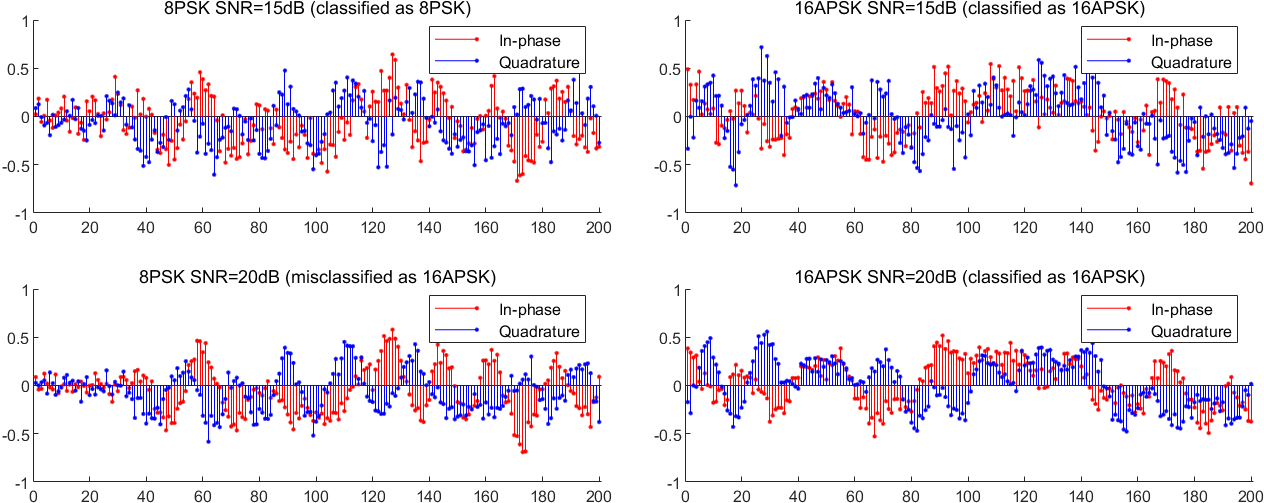}
\caption{Baseband signals in sequence representation}
\label{fig:robustness3}
\end{figure}

We compare our proposed distribution representation with image and sequence representations widely used in DL through some intuitive examples, as presented in Fig.~\ref{fig:robustness1} to Fig.~\ref{fig:robustness3}.

\noindent
\textbf{Signals as distributions in I/Q signal space}:
In Fig.~\ref{fig:robustness1}, four baseband signals of 8PSK \& 16APSK modulation formats at signal-to-noise ratio (SNR) of 15 \& 20 dB are depicted in distribution representation in I/Q space, of which their distributions are shown in heat maps, where yellow and blue regions indicate regions of high and low densities, respectively.

Two observations are that (a) the distributions of 8PSK and 16APSK signals at the same SNR (in the same row) are significant different, where the high density region for 8PSK is at the bottom-left quadrangle, and that of 16PSK is around the origin.  (b) The distributions of the signals sharing the same modulation format (in the same column) maintain a remarkable degree of similarity, at different  SNR = 15 \& 20 dB  in the presence of different levels of additive white Gaussian noise (AWGN).
It becomes more evident when we employ a distributional measure  to quantify the similarity between each pair of the distributions.
As illustrated in Fig.~\ref{fig:similarity}, the similarity between the distributions across different SNRs is higher when the modulation formats are identical than when they are different.

This example shows the effectiveness of distribution representation in capturing the essential features of signals with different modulation formats, as well as its robustness to noise.
This distribution representation enables our proposed classifier (described in Section~\ref{sec-IDK-OGD}) to correctly classify the modulation formats of these signals via the use of a distributional kernel.

\noindent
\textbf{Signals as images in polar coordinates}:
A recent image-based CNN classifier proposed by Teng \textit{et al.} \cite{teng2020accumulated} transforms I/Q data points in the original baseband signal sequence into a scatter plot of polar coordinates before converting it into an image with a resolution of $36 \times 36$ pixels.
The visual depictions of the same four baseband signals of 8PSK \& 16APSK modulation formats at SNR = 15 \& 20 dB are presented in Fig.~\ref{fig:robustness2}.
The difference between the images, representing signals of different modulation formats, is not pronounced. Neither is the similarity between the images of the same modulation format at different SNRs.
This shows that image representation falls short when compared to distribution representation, and it is not robust to noise.
It is interesting to note that CNN misclassifies the 8PSK signal at SNR = 20  dB (at a lower noise level) as 16APSK, when trained on 15 dB (at a higher noise level).

\noindent
\textbf{Signals in sequence representation}:
The fragments of the four baseband signal sequences in their original I/Q sequence representation are depicted in Fig.~\ref{fig:robustness3}.
While, to the naked eyes, the sequences for different formats can be differentiated relatively easily, even at different SNRs. However, the features extracted from a current state-of-the-art DL method might not provide such a distinction.
For example, the sequence-based AWN classifier \cite{zhang2023wavelet}, which utilizes the original I/Q baseband signal sequences as its input, misclassifies the 8PSK signal at SNR = 20 dB as 16APSK.
This is because the subsequent processing of I/Q baseband signal sequences within AWN for feature extraction is susceptible to the influence of noise.

It should be noted that unlike some DL methods which convert each set of I/Q points into an image with a certain resolution (e.g., \cite{peng2019constellation,wang2017constellation,mao2021attentive}), the proposed distribution representation eliminates the redundant step of the image conversion. This, in turn, avoids the loss of information as well as the introduction of noise.

In a nutshell, treating the constellation points in the I/Q signal space as a sample set from an unknown distribution is a representation of the original signals without information loss in AMC, which we show is better than existing representations employed in DL methods.
We provide a comparison with the sequence and image representations in the experiments reported in Section~\ref{sec-experiments}.

\subsection{Proposed signal representation using Isolation Distributional Kernel}
\label{sec-IDK}

Kernel methods have wide applications in classification and pattern recognition tasks. A point-to-point kernel measures the similarity between two points. Each kernel uniquely determines a feature mapping $\Phi$, which maps a point $x \in \mathbb{R}^n$ in the input space to a point $\Phi(x)$ 
 in the high-dimensional feature space called the Reproducing Kernel Hilbert Space (RKHS) \cite{scholkoph2001kernel}. 

Isolation Kernel (IK) is a recently proposed point-to-point kernel \cite{ting2018isolation,xu2019isolation}. The feature mapping of IK is built through random partitionings of an isolating mechanism. The parameters of an IK feature mapping include  $\psi$ and $t$, where $\psi$ is the number of partitions in each partitioning and $t$ is the number of partitionings. 

To be specific, let $D\subset \mathbb{R}^n$ be a given dataset. $\psi$ points $\{z_i\}$ ($i=1,2,\dots,\psi$) are randomly selected from $D$. Each point $z_i$ is isolated from the rest of the $\psi-1$ points by a hypersphere $\theta_i$, of which the radius $r_i$ is the distance between $z_i$ and its nearest neighbor among the $\psi-1$ points. 
By doing so, every point in the space belongs to either only one of these hyperspheres or none of them (if a point happens to be inside two or more hyperspheres, it belongs to the one which is closest to the point).

The feature map of IK, due to the partitioning $j$ of $\psi$ points, is defined as follows. Let $\phi_j(x|D)$ be a $\psi$-dimensional binary column vector defined by:
\begin{align}
\label{IK}
    \phi_j(x|D) = 
    \begin{cases}
        \mathbf{0} \quad \text{if} \quad d_{\min} > \underset{i\in[\psi]}{\max}\text{ }r_i\\
        [0,\cdots,0,1(k\text{th position)},0,\cdots,0]^\top \quad \text{otherwise}
    \end{cases}
\end{align}
where $d_{min}$ is the distance between $x$ and its nearest neighbor $z_k$ among the point set $\{z_i\}$ ($i=1,2,\dots,\psi$). 

Hereafter, `$|D$' is omitted when the context is clear. 

The complete feature mapping of IK $\phi: x\rightarrow \{0,1\}^{t\times\psi}$ is generated by repeating the above process $t$ times. For each $x \in D$, a $\psi$-dimensional vector $\phi_j(x)$ is created following (\ref{IK}) for partitioning $j$. The complete feature map $\Phi(x)$ is the concatenation of $\phi_1(x),\cdots,\phi_t(x)$, i.e.,
\begin{equation}
\label{IK2}
    \Phi(x) = [\phi_1(x)^\top,\cdots,\phi_t(x)^\top]^\top
\end{equation}

Based on IK, Isolation Distributional Kernel (IDK) is proposed to measure the similarity between two distributions \cite{ting2021-IDK}. The feature mapping of IDK, which maps the distribution $\mathcal{P}_S$ (which generates a set of points $S\subset \mathbb{R}^d$) to a point in the Hilbert space $\mathcal{H}$, is derived via the kernel mean embedding \cite{KernelMeanEmbedding2017} by using IK:
\begin{equation}
    \label{IDK_feature}
    \widehat{\Phi}(\mathcal{P}_S) = \frac{1}{|S|}\sum_{x\in S} \Phi(x)
\end{equation}
The similarity between two distributions $\mathcal{P}_S$ and $\mathcal{P}_T$, as measured by using IDK $\mathcal{K}$, is given as:
\begin{equation}
    \label{IDK_similarity}
    \mathcal{K}(\mathcal{P}_S, \mathcal{P}_T) = 
    \left< \widehat{\Phi}(\mathcal{P}_S), \widehat{\Phi}(\mathcal{P}_T) \right>
\end{equation}

We should point out that a conventional point-to-point kernel (e.g., Gaussian Kernel, Laplacian Kernel) can also lead to a distributional kernel via kernel mean embedding.  However, IK has two advantages that make it more suitable for the classification tasks like AMC than these conventional kernels :
\begin{itemize}
    \item The feature mapping of IK has finite dimensions, while both Gaussian Kernel and Laplacian Kernel have an infinite number of dimensions. A classifier cannot process data of infinite dimensionality. Though they can also have a finite-dimensional feature mapping via a kernel function approximation scheme such as Nystr{\"o}m method \cite{williams2000using} or Random Fourier Features \cite{rahimi2007random}, such approaches usually have high computational complexity and compromise the effectiveness of the original kernel function. 
    
    \item IK has a data-dependent feature mapping that derives directly from a given dataset. This data-dependent characteristic often leads to better task-specific performance. In contrast, conventional kernels with data-independent feature mappings tend to perform poorly in some tasks such as a Support Vector Machine (SVM) classification task \cite{ting2018isolation}.  
\end{itemize}

As a derivative of IK, IDK naturally inherits the above advantages. In addition, IDK also has the injective mapping guarantee which ensures that two distributions $\mathcal{P}_S$ and $\mathcal{P}_T$ are mapped to the same point in Hilbert space $\mathcal{H}$, i.e.,  \linebreak $\parallel \widehat{\Phi}(\mathcal{P}_S)- \widehat{\Phi}(\mathcal{P}_T) \parallel_\mathcal{H} = 0$, if and only if $\mathcal{P}_S=\mathcal{P}_T$ \cite{ting2018isolation}. Note that \textbf{none of the existing representations used in DL can make a similar guarantee}.  

In a nutshell, IDK is a natural choice to represent the high-level distributional features of I/Q baseband signals, and the representation can be performed at low cost without learning. The data-dependent mapping and the injective mapping guarantee are unique among existing representations. They are the key factors that lead to its high classification accuracy. Therefore, we choose IDK as the representation method for our proposed AMC scheme, to be described in the next section.

\section{Proposed online AMC scheme}
\label{sec-IDK-OGD}

We make use of a recent large scale online kernel classifier called Online Gradient Descent (OGD) which has been shown to work well in the context of conventional data streams \cite{OnlineKernelLearning-JMLR2016,Ting2021-X-Factor}. IK has been shown to be the key factor that enables OGD to run efficiently and effectively in a sequence of potentially infinite number of data points \cite{Ting2021-X-Factor}.

With the proposed IDK to represent I/Q baseband signals, applying OGD to the AMC context is simple, and it only requires the following modifications:

\begin{enumerate}
    \item Replace IK with IDK in IK-OGD \cite{Ting2021-X-Factor} in order to deal with I/Q baseband signals.
    
    \item Convert the two-class classifier to an $m$-class classifier (which consists of $m$ two-class classifiers) to deal with $m$ modulation formats.
    
    \item Modify the algorithm to have an explicit classification stage and a model-update stage for each small batch of incoming I/Q baseband signals in order to enable model update only if ground truths labels are available.
\end{enumerate}

Here we introduce the working principles of the proposed IDK-OGD classifier. A functional diagram is depicted in Fig.~\hyperref[fig:diagram]{\ref{fig:diagram}(b)}. Suppose there are $m$ different candidate modulation formats denoted as $M_1,M_2,\cdots,M_m$. We first need an initial batch of training data to construct $m$ distinct IDK feature mappings for each of the modulation formats. Namely, feature mapping $\widehat{\Phi}_j$ is constructed using the samples of modulation $M_j$ in the initial training data, based on equations (\ref{IK}) to (\ref{IDK_feature}). 

After every IDK-mapped point is obtained for each modulation format, all two-class OGD classifiers are used to produce a score and the final prediction is the one that has the maximum score. 

In the online setting, after a batch of samples (i.e., I/Q baseband signals) are classified by OGD classifiers, they are simultaneously updated only if the ground-truth modulation format labels of these samples are available.

\begin{algorithm}[!t]
\caption{IDK-OGD}
\label{IDK-OGD-pesudo}
\begin{algorithmic}[1]
\Require
    \Statex $m$ - The number of modulation formats
    \Statex $\widehat{\Phi}_j$  - IDK feature mapping of modulation format $j$
    \Statex $\eta$ - Learning rate
\Ensure
    \Statex Predicted modulation format $\widehat{c_i}$ for each signal $S_i \in \mathcal{B}$ (batch of signals)
\State Initialize weight vector $\mathbf{w}_j$ to $\mathbf{0}$ for $j=1,\dots,m$
\While{a new batch $\mathcal{B}$ arrives}
    \For{$i = 1 : |\mathcal{B}|$}
        \For{$j = 1 : m$}
            \State $g_{i,j} = \left< \mathbf{w}_j,\widehat{\Phi}_j(\mathcal{P}_{S_i}) \right >$ \Comment{score for format $j$}
        \EndFor
        \State $\widehat{c}_i = \arg\max_j \ g_{i,j}$
    \EndFor \Comment{Classification stage ends}
    \If{ground-truth labels $c_i$ for $S_i \in \mathcal{B}$ are available}
        \For{$i = 1 : |\mathcal{B}|$}
            \State $k_i=\begin{cases}$$1 &\mbox{ if }c_i=\widehat{c_i}$$\\$$-1&\mbox{ if }c_i\not=\widehat{c_i}$$\end{cases}$
            \For{$j = 1 : m$}
                \State $\mathbf{w}_j = \mathbf{w}_j - \eta\nabla L(g_{i,j};k_i)k_i\widehat{\Phi}_j(\mathcal{P}_{S_i})$
                \Statex \Comment{where $\nabla L$ is the gradient of loss function $L$}
            \EndFor
        \EndFor
    \EndIf \Comment{Online model update stage ends}
\EndWhile
\end{algorithmic}
\end{algorithm}

Algorithm~\ref{IDK-OGD-pesudo} shows the pseudo-code of the proposed IDK-OGD classifier which presents the technical details of classification and online updating stages of the proposed scheme. It is worth mentioning that since the calculations corresponding to each modulation format (i.e., the two for-loops in lines 4-6 and lines 12-14) are independent, the proposed IDK-OGD can be implemented using a parallel architecture, which can significantly enhance the efficiency.

Note that the distribution-based representation approach using IDK can be integrated with other algorithms besides OGD. In the experiment section, we have used IDK with Support Vector Machine (SVM) to create an IDK-SVM classifier for comparison.

\section{Experiments}
\label{sec-experiments}

\subsection{Experimental settings}

In this section, we design several experiments to evaluate the performance of the proposed IDK-OGD classifier under various time-varying channel conditions where there is a mismatch between the training data and testing data.

We compare the classification accuracy with fKNN as an online version of a feature-based moment KNN classifier \cite{zhu2015automatic} which uses the engineered high-order moment features as the representation of signals, a DL-based CNN classifier\footnote{The CNN classifier \cite{teng2020accumulated} is claimed to have an online version with a retraining process, but  the code is unavailable in the public domain. Only the no-retraining version of the CNN classifier is available, which is what we use here. In addition, we could not compare with recent methods  which (a) have claimed to have considered the effects of time-varying channel conditions (e.g., \cite{wang2021multi}), and (b) using constellation diagram images (e.g., \cite{peng2019constellation,mao2021attentive}), because their codes are  not available in the public domain.} with image representation that claimed to have achieved the state-of-the-art performance \cite{teng2020accumulated}, and a recent DL-based Adaptive Wavelet Network (AWN) classifier \cite{zhang2023wavelet} which adopts sequences for signal representation. In addition, we also compare with IDK-SVM, which uses the IDK features in SVM classification, to demonstrate the effectiveness of the proposed method under the time-varying channel conditions. The parameters used for all classifiers are shown in Table~\ref{tab:params}.

\setlength{\aboverulesep}{0pt}
\setlength{\belowrulesep}{0pt}
\begin{table}[t]
\renewcommand{\arraystretch}{1.5}
\caption{Experiment parameter settings}
\label{tab:params}
\begin{tabular}{c|c}
    \toprule
    Parameter & value \\
    \midrule
    Candidate modulation formats & \tabincell{c}{4ASK, 8ASK, BPSK, QPSK, 8PSK,\\16APSK, 32APSK, 16QAM, 32QAM, 64QAM}\\
    \midrule
    Length of each sample (signal) & $1024$\\
    \midrule
    $\psi$ for IDK & $128$\\
    \midrule
    $t$ for IDK & $75$\\
    \midrule
    Kernel for OGD and SVM & Linear\\
    \midrule
    Box constraint for SVM classifier & $980$\\
    \midrule
    $k$ for fKNN & $15$\\
    \midrule
    parameters for CNN & see \cite{teng2020accumulated}\\
    \midrule
    parameters for AWN & see \cite{zhang2023wavelet}\\
    \bottomrule
\end{tabular}
\end{table}

The experiments are conducted on the RadioML2018.01A dataset as well as synthetic datasets. RadioML2018.01A is an open-source dataset that includes both synthetic simulated channel effects and over-the-air recordings of 24 digital and analog modulation formats \cite{timothy2016RadioML}. It considers more realistic channel impairments (compared to the earlier released RadioML2016.10A and RadioML2016.10B datasets) such as carrier frequency offset, multipath fading, sample rate offset, etc., which make the classification task more challenging. Note that the RadioML2018.01A dataset has only SNR labelled, and other channel impairments mentioned above are not given in the dataset. Therefore, we create a series of synthetic datasets using MATLAB Communications Toolbox to examine different simulated channel effects.

In the experiments, all algorithms use the same initial training set to obtain the initial classifiers. As new samples come in batches, each algorithm has its own way to make use of the new batches. IDK-OGD is the only one which updates its classifier incrementally with every new batch of data, after classifying all the samples in the batch (as shown in Algorithm~\ref{IDK-OGD-pesudo}). fKNN combines every new batch with the previous training set to retrain its classifier. IDK-SVM, CNN and AWN always use the same initial classifier to make predictions, and do not retrain using the new batches since they have no incremental model update mechanism, and it would be too time-consuming for these batch learning algorithms to retrain for every new batch using a training set which combines all the previous training data and the new batch \cite{teng2020accumulated, zhang2023wavelet}. The model retraining or update mechanisms of the five algorithms are summarized in Table~\ref{tab:training}.

\begin{table}[t]
\renewcommand{\arraystretch}{1.5}
\caption{Retraining or model update used in five algorithms}
\label{tab:training}
\begin{tabular}{c|c|c|c|c}
    \toprule
    IDK-OGD & fKNN & IDK-SVM & CNN & AWN\\
    \midrule
    Online update & Online retrain & \multicolumn{3}{c}{No retrain}\\
    \bottomrule
\end{tabular}
\end{table}

\begin{sidewaystable}[!p]
\renewcommand{\arraystretch}{1.5}
\caption{Experimental settings of experiments under mismatched channel conditions in training and testing sets}
\label{tab:settings_mismatched}
\begin{tabular}{c|c|c|c|c|c|c|c}
    \toprule
    \multirow{2}{*}[-1.5em]{Dataset} & \multirow{2}{*}[-0.5em]{\tabincell{c}{Mismatched channel\\impairments of\\training and testing sets}} & \multirow{2}{*}[-1.5em]{\tabincell{c}{Experiment\\identifier}} & \multicolumn{2}{|c|}{Training} & \multicolumn{3}{|c}{Testing}\\
    \cmidrule{4-8}
    & & & \tabincell{c}{Dataset\\size} & \tabincell{c}{Channel\\conditions} & \tabincell{c}{Dataset\\size} & \tabincell{c}{Number of\\batches\\per lot} & \tabincell{c}{Channel conditions\\for one lot}\\
    \midrule
    \multirow{2}{*}[-1em]{\tabincell{c}{RadioML\\2018.01A}} & \multirow{2}{*}[-1em]{SNR with AWGN} & $\mathbb{R}_u$ & \multirow{2}{*}[-1em]{\tabincell{c}{$10000$\\samples}} & \tabincell{c}{SNR in $[0,6]$ dB\\with step = $2$ dB} & \multirow{2}{*}[-1em]{\tabincell{c}{$110000$\\samples}} & \multirow{2}{*}[-1em]{$1$} & \multirow{2}{*}[-1em]{\tabincell{c}{Some SNR\\between $[14,20]$ dB}}\\
    \cmidrule{3-3} \cmidrule{5-5}
    & & $\mathbb{R}_o$ & & \tabincell{c}{SNR in $[6,14]$ dB\\with step = $2$ dB} & & & \\
    \midrule
    \multirow{4}{*}[-4em]{Synthetic} & SNR with AWGN & $\mathbb{S}_u^1$ & \multirow{4}{*}[-4em]{\tabincell{c}{$5000$\\samples}} & \multirow{4}{*}[-4em]{\tabincell{c}{SNR = $15$ dB,\\phase noise\\=$-9999$ dBc/Hz,\\I/Q a.i. = $0$ dB}} & \multirow{4}{*}[-4em]{\tabincell{c}{$11000$\\samples}} & \multirow{4}{*}[-4em]{$10$} & \tabincell{c}{Some SNR\\between $[10,20]$ dB}\\
    \cmidrule{2-3} \cmidrule{8-8}
    & Phase noise & $\mathbb{S}_u^2$ & & & & & \tabincell{c}{Some phase noise\\between $[-40,-30]$ dBc/Hz}\\
    \cmidrule{2-3} \cmidrule{8-8}
    & \tabincell{c}{I/Q amplitude\\imbalance} & $\mathbb{S}_u^3$ & & & & & \tabincell{c}{Some I/Q a.i.\\between $[-5,5]$ dB}\\
    \cmidrule{2-3} \cmidrule{8-8}
    & \tabincell{c}{All 3 types of\\channel impairments} & $\mathbb{S}_u^4$ & & & & & \tabincell{c}{SNR: $[10,15]$ dB,\\phase noise:\\$[-40,-30]$ dBc/Hz,\\I/Q a.i.: $[-5,5]$ dB}\\
    \bottomrule
\end{tabular}
\footnotetext{Limited matched conditions are also included.}
\end{sidewaystable}

Note that most existing methods are based on the assumption that training and testing data are under the same and fixed channel conditions. But in real-world AMC applications, it is impractical to estimate SNR or other channel parameters since the estimations are often inaccurate. Therefore, we do not make such assumption, and it is necessary to conduct experiments to evaluate the performance of the classifiers using testing sets of which the channel conditions mismatch those of the training sets.

The first two experiments are conducted on the RadioML2018.01A dataset. In both of these experiments, the testing sets remained consistent and have high SNR, while the training sets exhibited low SNR in one case and medium SNR in the other, with the SNR range of the latter overlapping with that of the testing set.

Furthermore, we design four other experiments with the training and testing sets of mismatched channel conditions and different channel impairments, including additive white Gaussian noise (AWGN), phase noise, I/Q amplitude imbalance, and a combination of all the three impairments. Since the only parameter we can control in the RadioML2018.01A dataset is the SNR, these 4 experiments are conducted on our synthetic datasets to study the performance of IDK-OGD under time-varying channel conditions with the training and testing data having different channel impairments. The detailed settings for the training and testing sets of the experiments are given in Table~\ref{tab:settings_mismatched}.

Stratified sampling is employed to generate samples in such a way that each modulation format is represented by an equal number of samples within every batch in the testing set. Samples in the testing set come in batches, each containing $100$ samples, and the batches are input into the classifier one by one to mimic real-time arrival. To simulate time-varying channel conditions, we group a specific number of batches together (designated in Table~\ref{tab:settings_mismatched}), forming a single ``lot''. Each lot consists of multiple batches that share identical channel conditions, while different lots may exhibit different channel conditions.

\subsection{Experimental results}

\begin{figure}[t]
\centering
\includegraphics[width=0.95\textwidth]{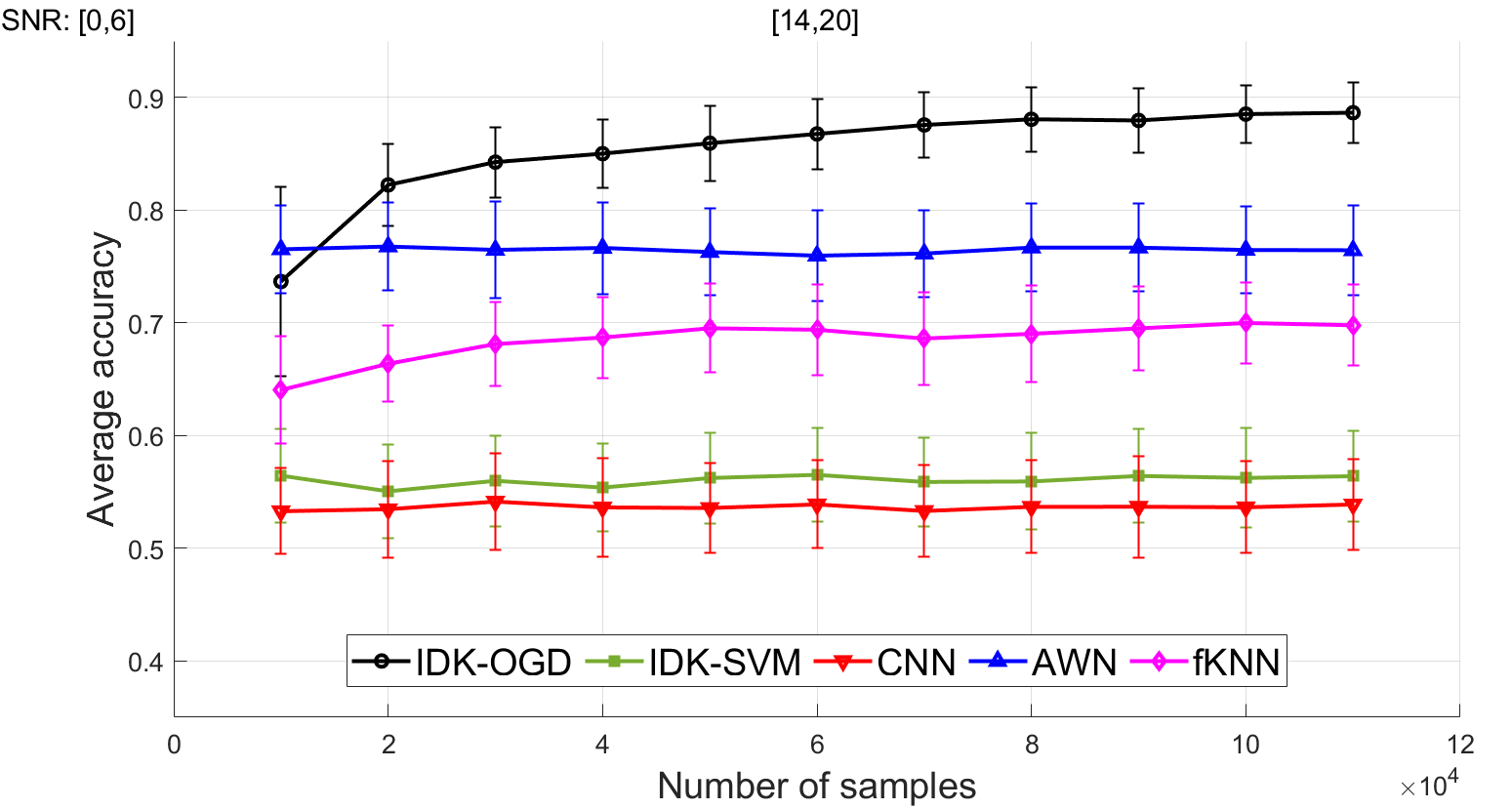}
\caption{Experimental results of experiment id: $\mathbb{R}_u$ (shown in Table~\ref{tab:settings_mismatched}), where training and testing SNR ranges are $[0, 6]$ dB and $[10, 20]$ dB, respectively, as stated in Table~\ref{tab:settings_mismatched}}
\label{fig:mismatch_RadioML}
\end{figure}

\begin{figure}[t]
\centering
\includegraphics[width=0.95\textwidth]{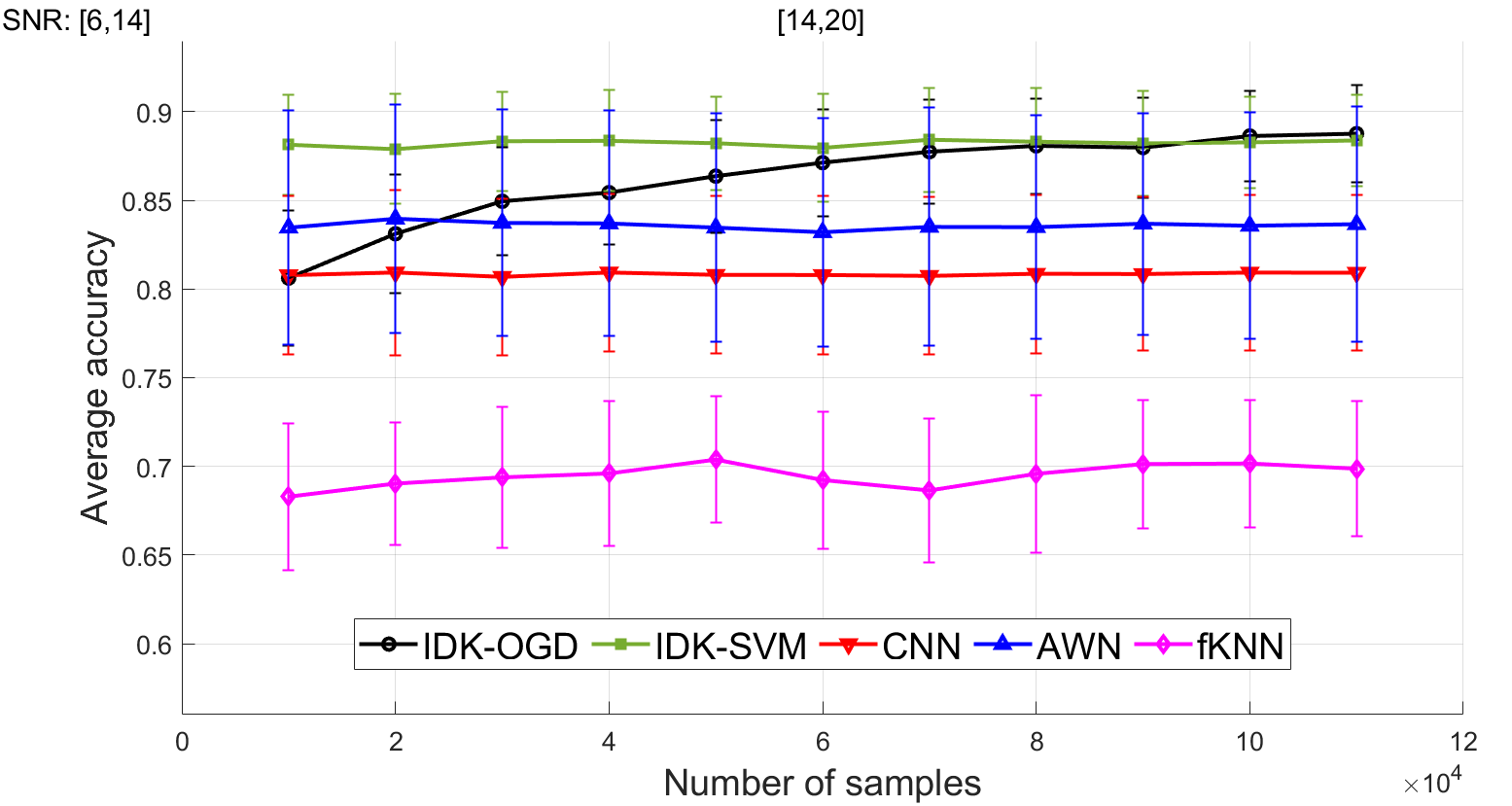}
\caption{Experimental results of $\mathbb{R}_o$}
\label{fig:overlap_RadioML}
\end{figure}

\begin{figure}[t]
\centering
\includegraphics[width=0.95\textwidth]{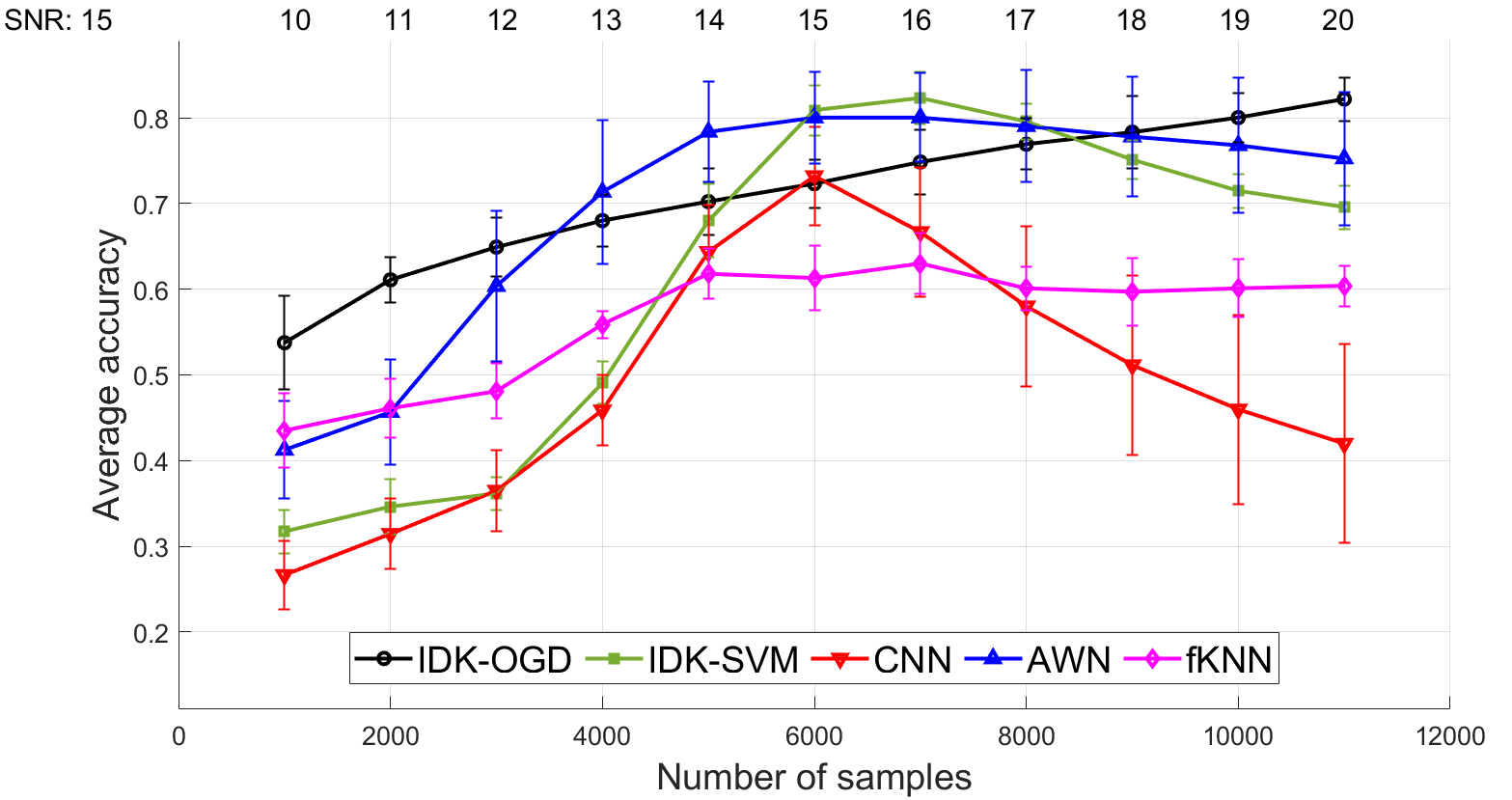}
\caption{Experimental results of $\mathbb{S}_u^1$}
\label{fig:mismatch_snr}
\end{figure}

\begin{figure}[t]
\centering
\includegraphics[width=0.95\textwidth]{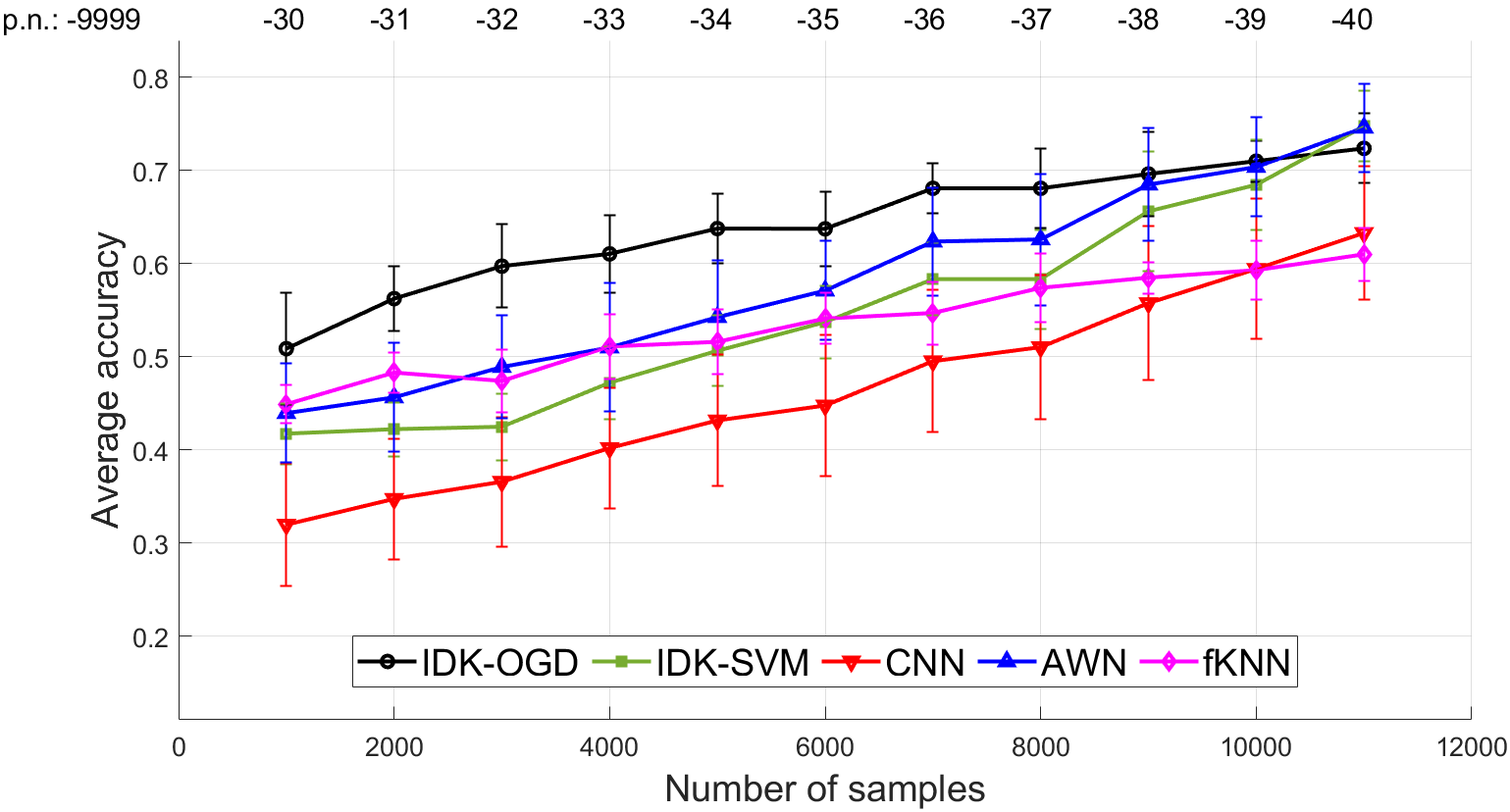}
\caption{Experimental results of $\mathbb{S}_u^2$}
\label{fig:mismatch_pn}
\end{figure}

\begin{figure}[t]
\centering
\includegraphics[width=0.95\textwidth]{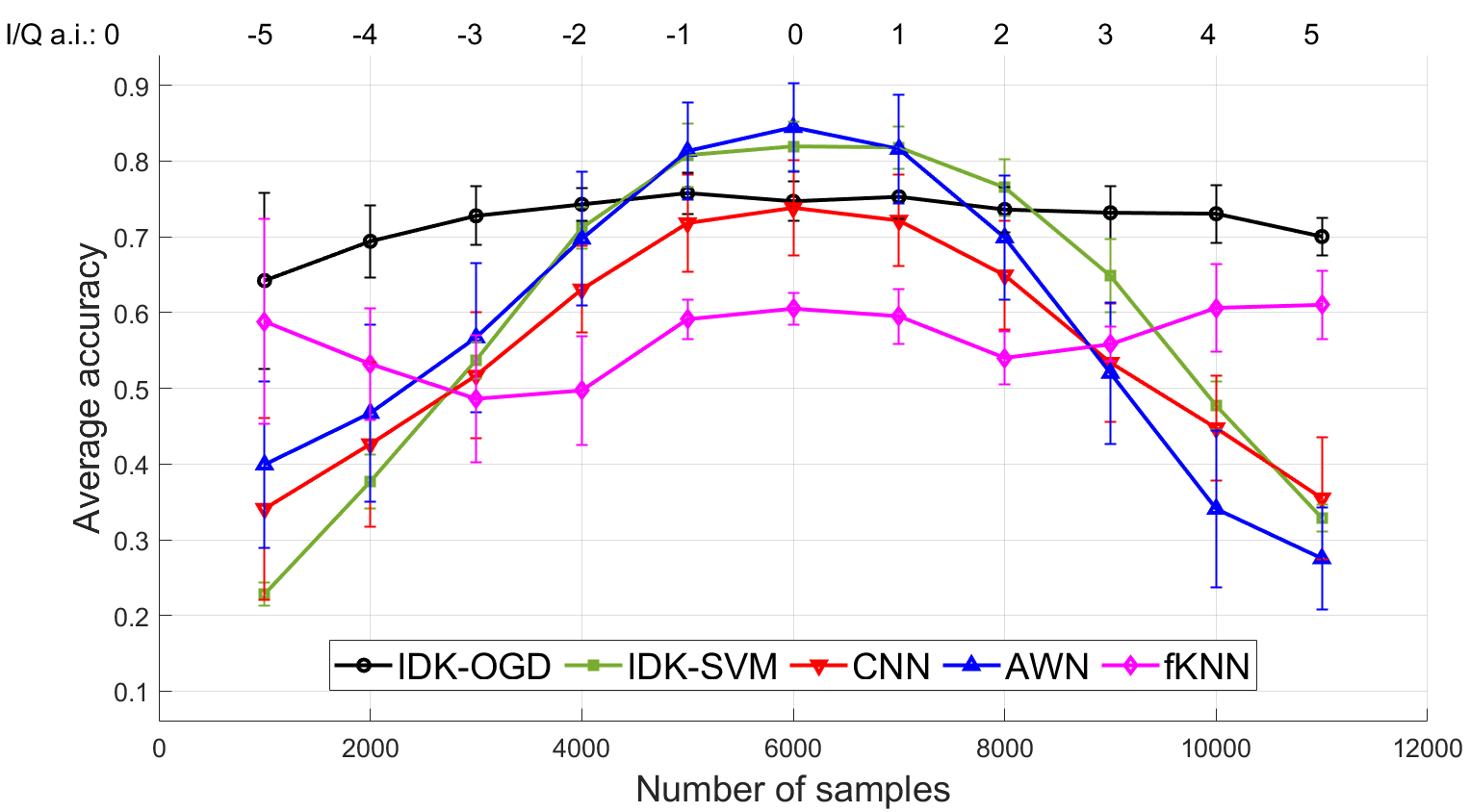}
\caption{Experimental results of $\mathbb{S}_u^3$}
\label{fig:mismatch_iq}
\end{figure}

\begin{figure}[t]
\centering
\includegraphics[width=0.95\textwidth]{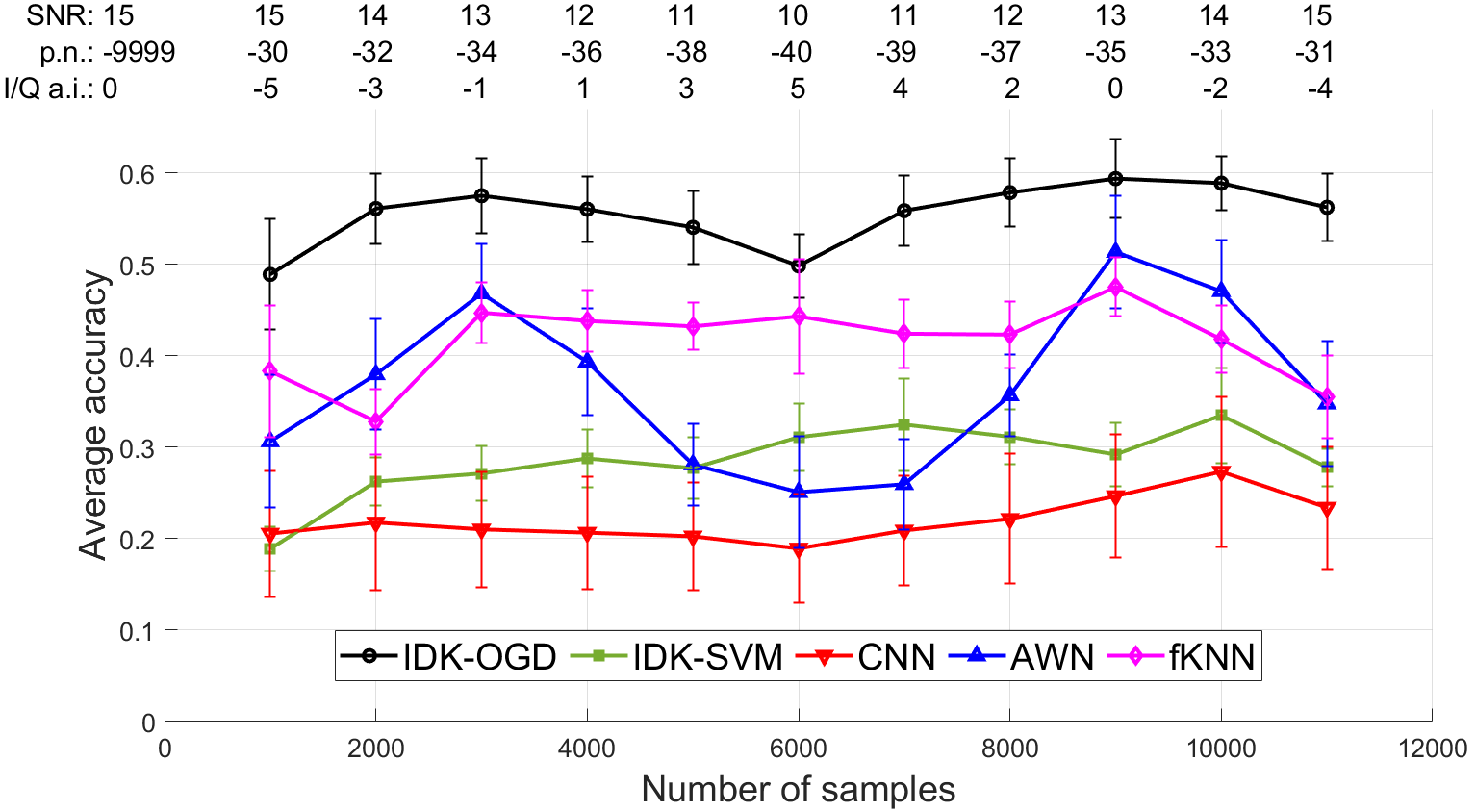}
\caption{Experimental results of $\mathbb{S}_u^4$}
\label{fig:mismatch_all}
\end{figure}

The experimental results are shown in Fig.~\ref{fig:mismatch_RadioML} to Fig.~\ref{fig:mismatch_all} where the performance of each classifier is measured in terms of prediction accuracy. Each average accuracy and its standard error bars are taken over every $100$ batches (for $\mathbb{R}_u$ and $\mathbb{R}_o$ which are conducted on the RadioML2018.01A dataset) or $10$ batches (for the other four experiments conducted on synthetic datasets) and over $10$ trials. The top row of the figure displays the channel conditions within the training set (top left), along with the varied channel conditions over time in the testing set (the remaining values in the top row), which are directly associated with the average accuracies presented below them.

Our main observations are presented as follows.

\begin{itemize}
    \item Our proposed IDK-OGD is almost always the best classifier among the five classifiers on the RadioML2018.01A dataset when there is a mismatch between the SNRs of the training and testing samples, as can be seen in Fig.~\ref{fig:mismatch_RadioML}.  
    
    When the SNR range of the training set overlaps with that of the testing set, as shown in Fig.~\ref{fig:overlap_RadioML}, IDK-OGD makes use of the training data to update its model and becomes the best classifier during the process, which is consistent with the results of $\mathbb{R}_u$ in Fig.~\ref{fig:mismatch_RadioML}.

    Note that, in both Fig.~\ref{fig:mismatch_RadioML} and Fig.~\ref{fig:overlap_RadioML},  IDK-OGD performs slightly worse than AWN at the beginning of the experiment, but thanks to its capability of online model update, IDK-OGD utilizes the new batches to update the classifier to quickly adapt to the channel conditions of the testing set, outperforming AWN and becoming the best classifier ever since.
    
    Fig.~\ref{fig:mismatch_snr} to Fig.~\ref{fig:mismatch_iq} show the comparison results which have both matched and mismatched conditions in terms of SNR, phase noise and I/Q amplitude imbalance. They all demonstrate that when the simulated channel conditions of the testing set mismatch those of the training set, IDK-OGD performs better than the other four classifiers. And when all the three impairments are present, as depicted in Fig.~\ref{fig:mismatch_all}, the performance of all the other classifiers drops significantly, but IDK-OGD still outperforms the other four classifiers by a large margin.
    
    \item IDK-SVM, CNN and AWN have similar accuracies and behaviors under many time-varying channel conditions, i.e., they have better accuracies than IDK-OGD if and only if the training and testing data have similar channel conditions; otherwise, their accuracy are degraded. The lack of updating or retraining mechanism sets the accuracy limit they can achieve.
    
    AWN exploits the features in different frequency bands of the signal via adaptive wavelet decomposition, making it more robust to the channel impairments than IDK-SVM and CNN and thus achieving a higher accuracy most of the time. But AWN cannot update its model to adapt to the changing channel conditions, and its accuracy drops significantly when the testing data have different channel conditions from the training data, and AWN is no match for IDK-OGD in this case.

    \item fKNN exhibits robustness against the mismatch channel conditions of the training and testing sets, consequently resulting in better performance in comparison to IDK-SVM and CNN under such circumstances, particularly when the mismatch is substantial. However, despite the fact that fKNN has an online retraining process, it still often performs worse than AWN, and its performance is always much worse than IDK-OGD.
\end{itemize}

\begin{figure}[!p]
\centering
\begin{minipage}[b]{\textwidth}
    \centering
    \includegraphics[width=\textwidth]{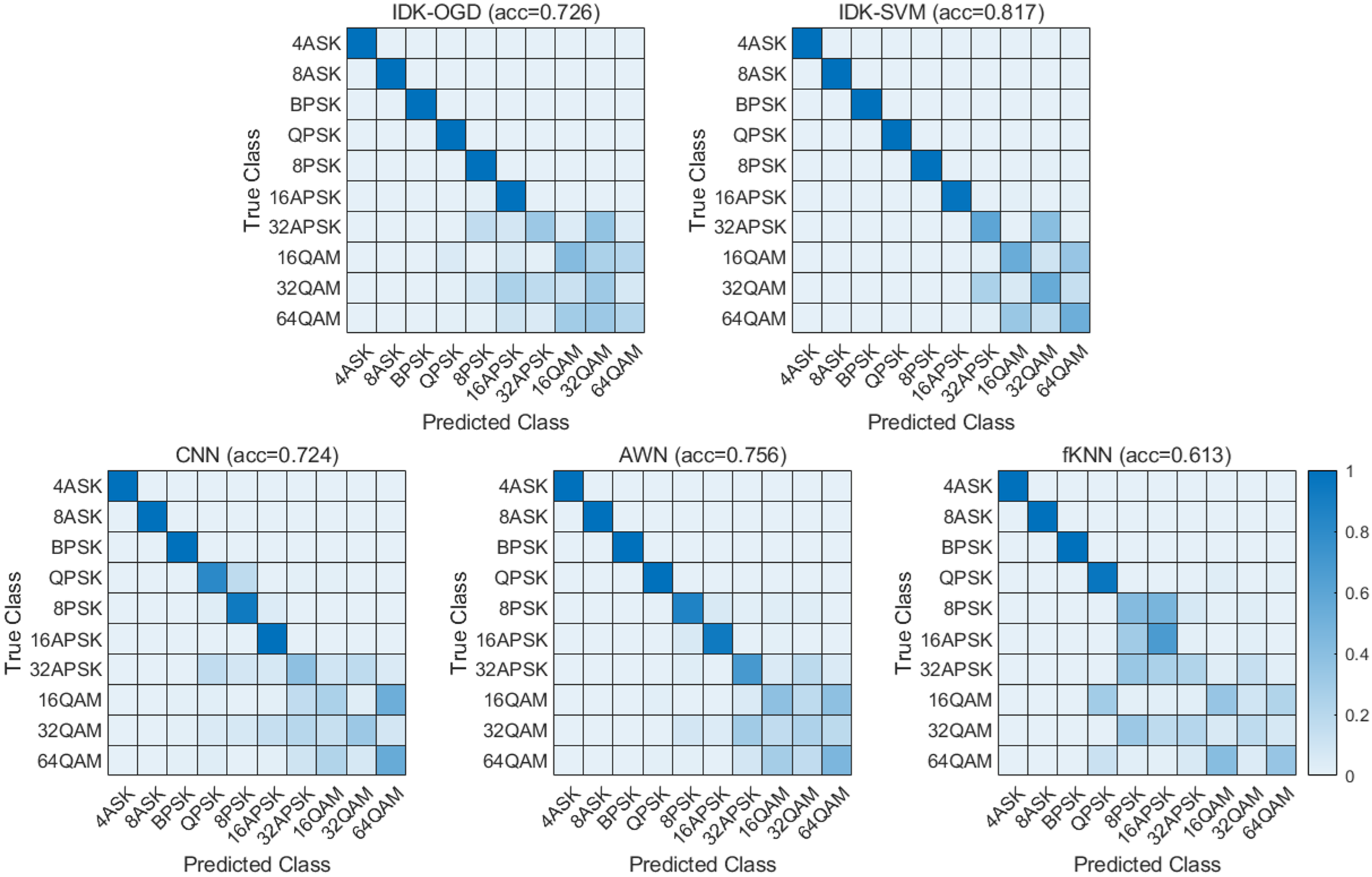}\par
    {\footnotesize\textbf{(a)} SNR = 15 dB}
\end{minipage}
\hfill
\vspace{0pt}
\begin{minipage}[b]{\textwidth}
    \centering
    \includegraphics[width=\textwidth]{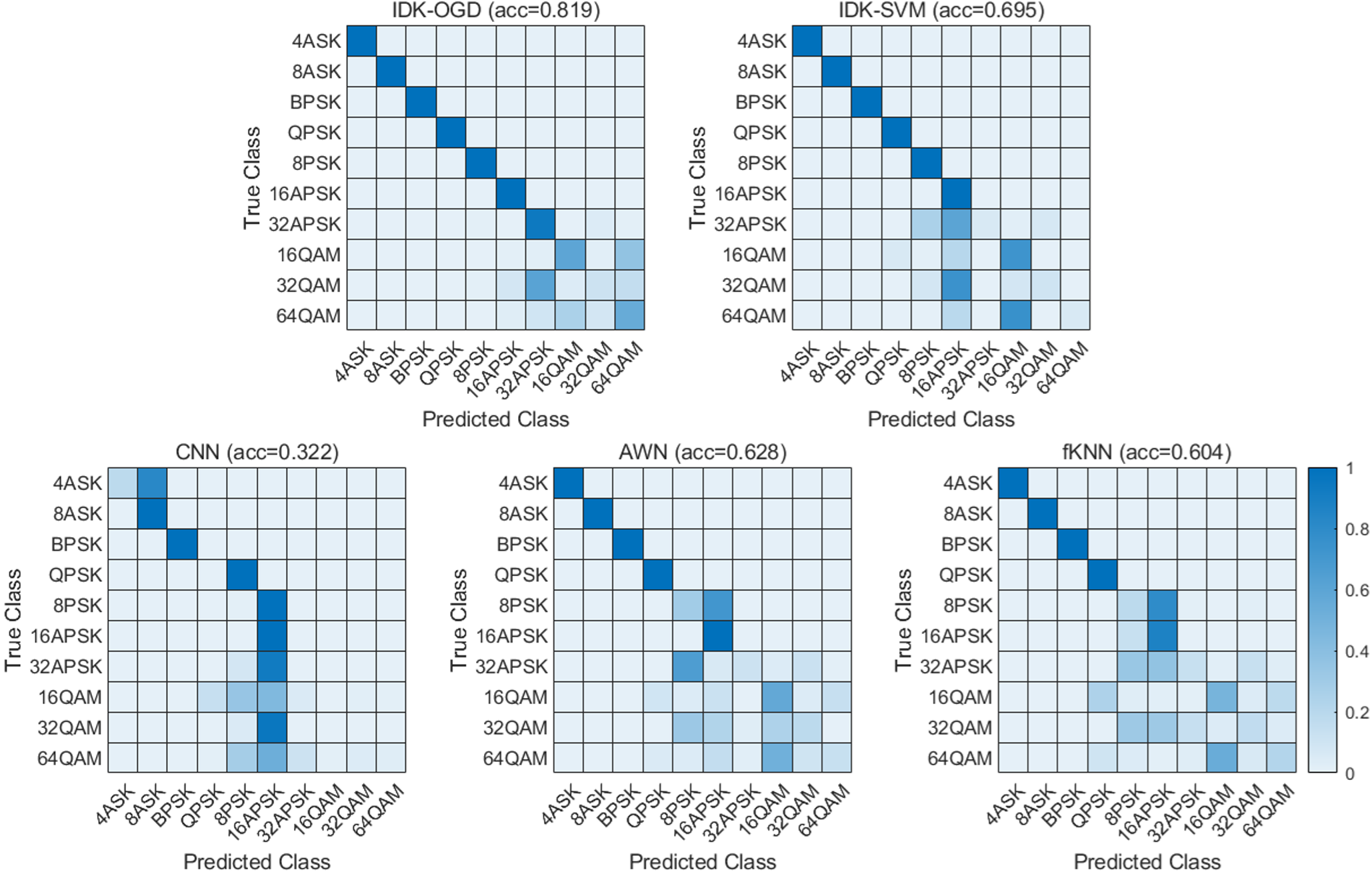}\par
    {\footnotesize\textbf{(b)} SNR = 20 dB}
\end{minipage}
\caption{Confusion matrices for all classifiers in $\mathbb{S}_u^1$ at an testing SNR of (a) 15 dB and (b) 20 dB. In both cases, each classifier is trained on a training set with an SNR of 15 dB}
\label{fig:confs}
\end{figure}

Fig.\ref{fig:confs} illustrates the confusion matrices of all five classifiers in $\mathbb{S}_u^1$, which reveal the classification errors between different modulation formats for each classifier. The confusion matrices in Fig.~\hyperref[fig:confs]{\ref{fig:confs}(a)} correspond to the results at SNR = 15 dB, while those in Fig.~\hyperref[fig:confs]{\ref{fig:confs}(b)} correspond to the results at SNR = 20 dB. Each classifier trained at SNR = 15 dB is used for testing in both cases. 

The results show that all classifiers have high errors in predicting 32APSK, 16QAM, 32QAM \& 64QAM under the matched channel condition of SNR = 15 dB (shown in Fig.~\hyperref[fig:confs]{\ref{fig:confs}(a)}).

As the testing SNR increases to 20 dB (i.e. with a lower noise level than the classifiers are trained from), the accuracies of all classifiers, except IDK-OGD,  deteriorate markedly due to the mismatched channel condition (shown in Fig.~\hyperref[fig:confs]{\ref{fig:confs}(b)}). For example, AWN often makes mistakes between 16QAM and 64QAM at the matched SNR = 15 dB.
However, when the SNR increases to 20 dB, AWN tends to misclassify 8PSK as 16APSK and 32APSK as 8PSK, which aligns with the classification results of AWN in Fig.~\ref{fig:robustness3}. This observation shows that under mismatched channel conditions, AWN loses its ability to accurately classify some of the modulation formats because it is incapable of adjusting to varying channel conditions.

In contrast, IDK-OGD maintains a high level of prediction accuracy across almost all modulation formats, even under the mismatched condition. This exceptional performance can be attributed to the utilization of new signal batches with reduced noise at an SNR of 20 dB, enabling the model to undergo essential updates and, consequently, achieve higher accuracy. Note that IDK-OGD is the only classifier which improves its accuracy when noise level reduces from SNR = 15 dB to 20 dB.

A comparison between IDK-OGD and IDK-SVM shows the importance of the adaptive model update of OGD. Though they both employ the same IDK for signal representation, IDK-SVM achieves significantly lower accuracy because it does not update its model. We postulate that the poor accuracies of AWN and CNN are mainly due to the weaker representations of sequence and image, respectively, though no model update also plays a part. The fKNN classifier is not competitive because the engineered features are not good enough, even under matched conditions.

In summary, the experimental results show that the proposed IDK-OGD is the only classifier that works well in online settings under mismatched channel conditions. Two key factors contribute to this outcome, i.e., a powerful distributional kernel is used for representation of I/Q baseband signals, and the effective adaptive model update of OGD.

\section{Ablation studies}
\label{sec-ablationStudy}

We conduct two ablation studies to examine (1) the effect of different SNR ranges and sizes of the training sets; and (2) the relative accuracy of IDK and the representation learning in DL methods (i.e., CNN and AWN).

\subsection{The effect of different SNR ranges and sizes of the training sets}

\begin{figure}[t]
\centering
\includegraphics[width=\textwidth]{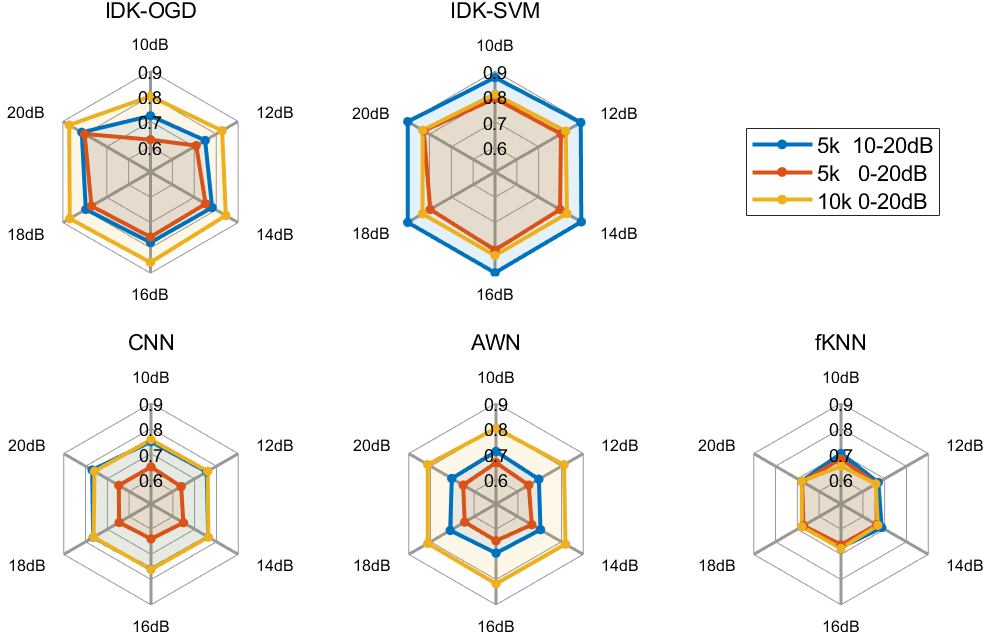}
\caption{The average test accuracy of the classifiers trained on training sets with different SNR ranges and sizes (shown in the legend) on the RadioML2018.01A dataset. The testing SNR range is $[10, 20]$ with a step of 2 dB in all cases}
\label{fig:range}
\end{figure}

Under mismatched channel conditions, some of the channel impairments in the testing set differ from those in the training set. This makes the classification task more challenging. We show here that one cannot simply use a training set with signals of all possible channel conditions to train a classifier since it would degrade the performance of the classifier. 

Two comparisons are conducted here. (i) Different SNR ranges of training sets: $[10, 20]$ dB versus $[0, 20]$ dB having 5,000 training samples. (ii) Different training set sizes: 5,000 versus 10,000 training samples with SNR in $[0, 20]$ dB.

The first comparison result is demonstrated in Fig.~\ref{fig:range} between the blue and red lines.
It indicates that the classifiers trained on the training set with an SNR range of $[0, 20]$ dB always perform worse than those trained on the training set with an SNR range of $[10, 20]$ dB. This shows that incorporating an excessive number of channel impairments into the training set may not be a judicious course of action.

The second comparison is shown in Fig.~\ref{fig:range} between the red and yellow lines. Here we observe that, when the SNR ranges of the training sets are identical ($[0, 20]$ dB), a larger training set size leads to a better performance.

It is noteworthy that IDK-SVM appears to perform the best when trained on the training set with SNR in $[10, 20]$ dB. This is because IDK-SVM is able to achieve a higher accuracy than IDK-OGD when the SNR ranges of the training and testing sets are exactly the same. Otherwise, IDK-OGD outperforms IDK-SVM, as evidenced by the results of $\mathbb{R}_u$ in Fig.~\ref{fig:mismatch_RadioML} and $\mathbb{R}_o$ in Fig.~\ref{fig:overlap_RadioML}.

\subsection{IDK vs the representation learning in DL methods}

\begin{figure}[t]
\centering
\includegraphics[width=0.95\textwidth]{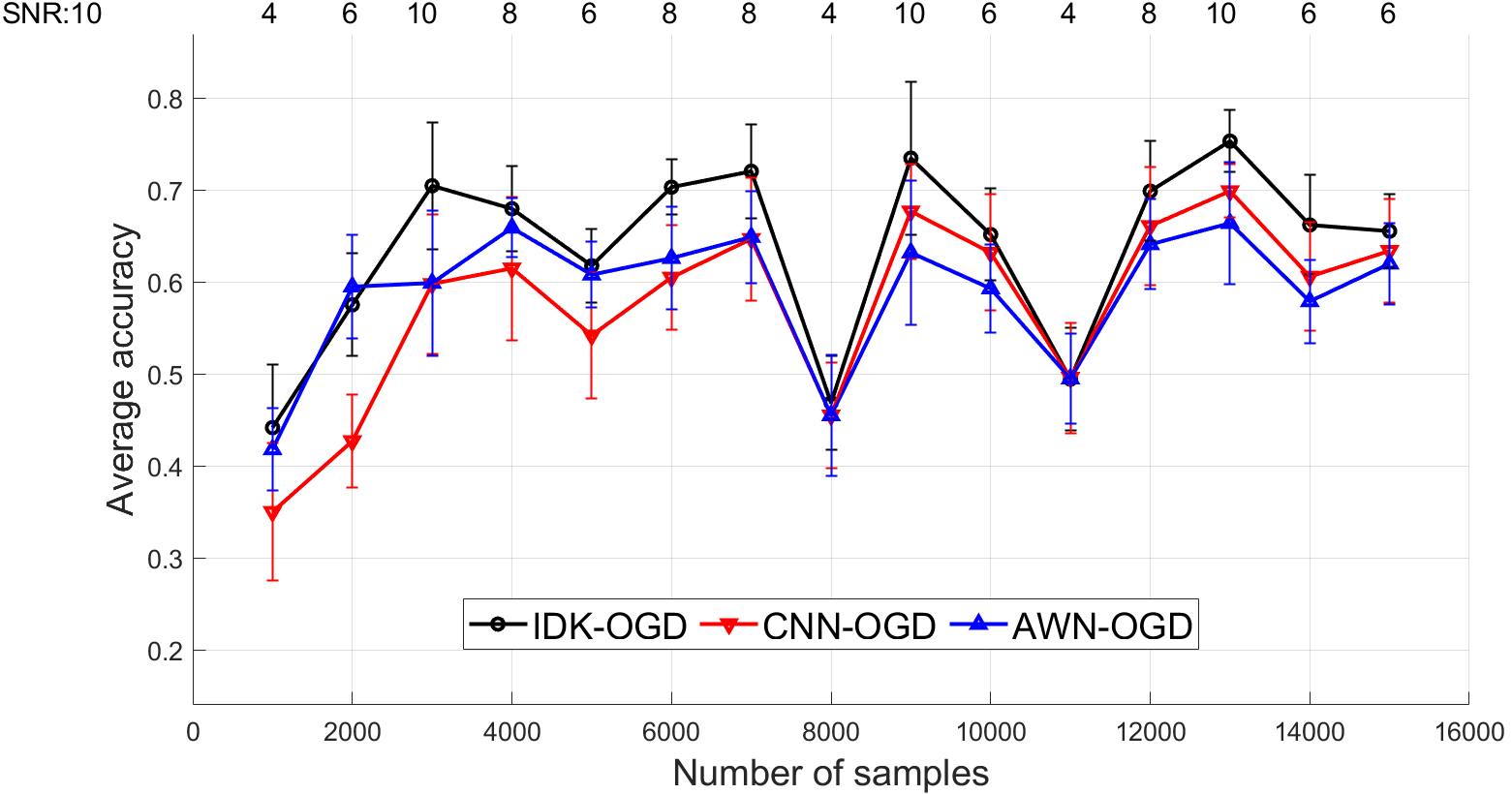}
\caption{The accuracies of IDK-OGD, CNN-OGD and AWN-OGD}
\label{fig:rps}
\end{figure}

Here we show the superiority of IDK over the representation learning in the two DL classifiers via an experiment where we compare the performance of IDK-OGD with CNN-OGD and AWN-OGD. CNN-OGD utilizes the output of the pooling layer (i.e., the result of the representation learning) in CNN, while AWN-OGD employs the computed features prior to the final fully connected layers (i.e., before classification) in AWN, for OGD classification. The result given in Fig.~\ref{fig:rps} shows that IDK-OGD almost always outperforms both CNN-OGD and AWN-OGD. This shows the power of the no-learning distribution representation of IDK and its advantage over the non-distribution-based representation via the two deep learning methods (using the sequence and image representations).

\begin{figure}[t]
\centering
\includegraphics[width=0.95\textwidth]{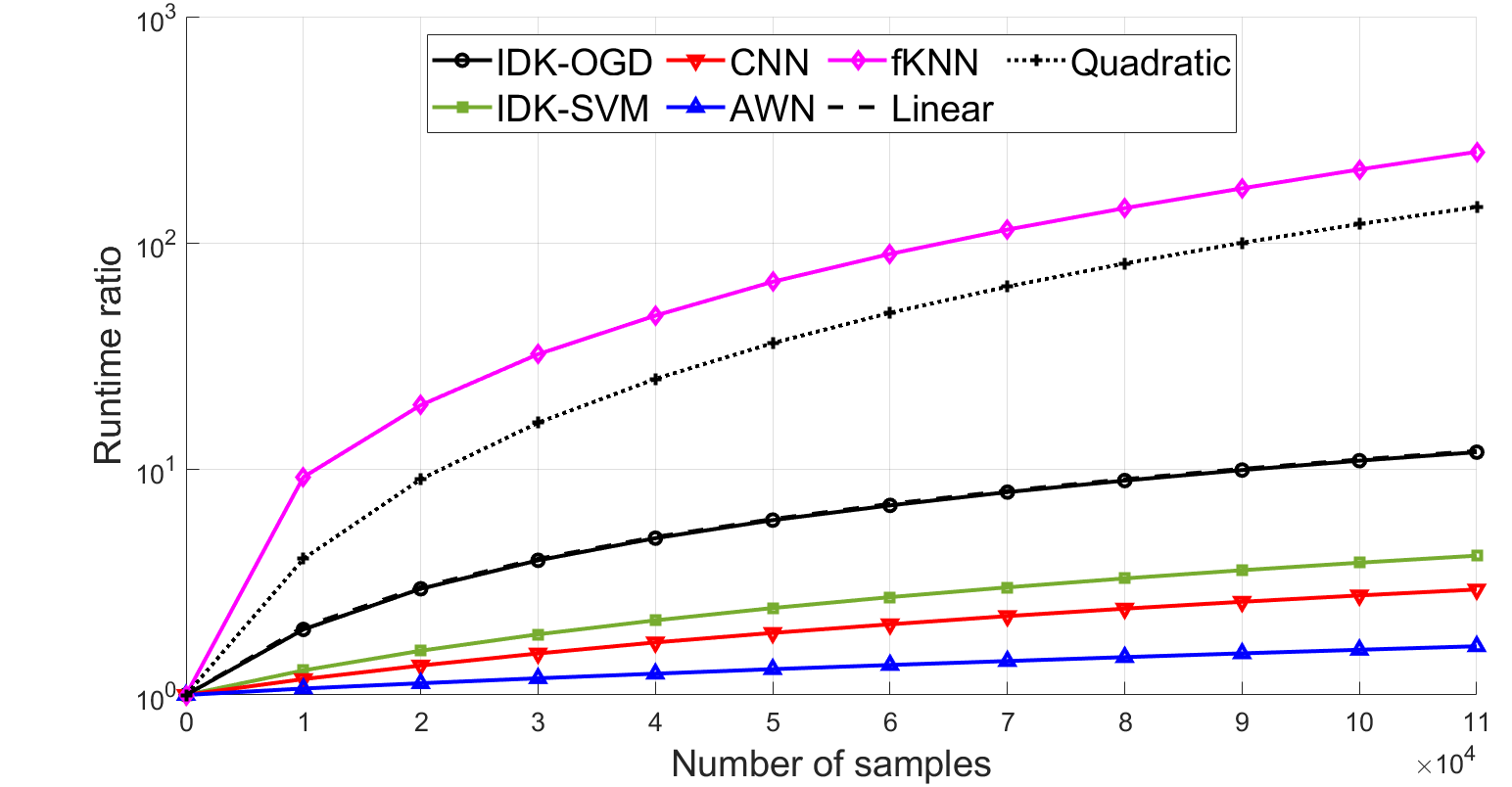}
\caption{Runtime ratio for the classifiers. The IDK-OGD curve overlaps with the linear curve}
\label{fig:time_ratio}
\end{figure}

\section{Runtime comparison}
\label{sec-runtime}

Fig.~\ref{fig:time_ratio} shows the ratio of the total runtime to the training time. IDK-OGD has its runtime growing linearly with the number of samples in the testing set, while fKNN has quadratic time complexity. The runtime ratios of IDK-SVM, CNN and AWN are lower than IDK-OGD because they are batch learning algorithms without a retraining process. Note that, despite their fast predicting time in batch mode,  it is still impractical to perform retraining on a combined training set (as in fKNN). This is because every retraining would be too time-consuming with the tuning of their parameters.

In fact, the time complexity of IDK-OGD is linear to the input data size $n$, i.e., $O(nmLt\psi)$. Here, $m$ denotes the number of candidate modulation formats, $L$ is the length of each signal sequence, while both $t$ and $\psi$ are the parameters of IDK, and all of them are constants that remain fixed and entirely independent of the input data size $n$.

In summary, our proposed IDK-OGD classifier is the only online learning algorithm with a model update having linear time complexity, as far as known.

\section{Discussion}
\label{sec-discussion}

We show that the distribution representation using IDK is better than existing representations of signals currently used in deep learning. A key determinant of this outcome (not discussed thus far) is IDK's data dependency property: two distributions are more similar to each other when measured by IDK derived from a sparse region than that from a dense region (see the details in \cite{ting2021-IDK}).

Indeed,  recent comparisons in other tasks  have shown that deep learning often does worse than no-learning feature representation methods (e.g., in time series \cite{FlawedBenchmarks2021,ScalableSubsequenceAnomalyDetection-VLDB2021,Rocket-2020,schmidl2022anomaly}). Our result is consistent with these results.

The complexity of a deep learning classifier (e.g., architecture, many tuning parameters, large training set size requirement) has been a key obstacle in making it applicable in the online setting. We are not aware of the availability of an online deep learner in AMC, despite the claim of some papers (e.g., \cite{teng2020accumulated}) as no codes are made available allowing others to repeat their experiments. The same applies to \cite{mao2021attentive}, though they have conducted an extensive evaluation using the RadioML2018.01A dataset with matched SNR in the training and testing sets.

Looking forward, there are ways to improve upon IDK-OGD. First, it is possible to further improve the data dependency of IDK by re-initializing it during testing. The reported experiments have initialized IDK at the initial training stage only. Second, improving the robustness to corruptions to the signals (in the transmission process) in OGD will further increase its classification accuracy. These corruptions are the mismatched conditions between the training set and testing set we have simulated using the RadioML2018.01A and synthetic datasets. Third, it is worth contemplating the adoption of a more complex classifier than OGD to improve the classification accuracy under matched channel conditions.

The insight has a much wider implication. First, existing as well as any new distribution measures can now be used in AMC.
Second, can deep learning add any value to the powerful no-learning distribution representation? We show that the existing non-distribution representations used in DL, despite having additional learning, work worse than the no-learning distribution representation. Third, is a complicated classifier warranted, given the use of the powerful distribution representation? We show that a simple OGD is more robust than DL in various conditions, with the help of the distribution representation. These are interesting issues for future investigation.

\section{Conclusions}
\label{sec-conclusion}

We provide the insight that a constellation diagram of baseband signals can be intuitively and effectively represented
as a distribution, without loss of information. It is
exactly because of the lack of this insight that no one
has used a distribution measure in this way in AMC
before.

This insight empowers a distribution measure to be used in AMC to both represent baseband signals and
perform similarity measurement between any two
sets of baseband signals. 

With this insight, we show that a recent distributional kernel working in concert with an existing online kernel classifier becomes a powerful method in AMC, where no credible existing methods have been shown to work in online settings before. 

We show for the first time that an online AMC scheme with a distribution representation of signals works well under realistic time-varying channel conditions.  The proposed method called IDK-OGD has two parts: First, the distributional kernel IDK is used to map each baseband signal, as a set of i.i.d. two-dimensional points of an unknown distribution in the I/Q domain, to a point in the Hilbert space of IDK. Second, OGD, as an online kernel classifier, has been shown to work well in point-based data streams. We show through extensive experiments that the proposed IDK-OGD classifier is most effective as an online AMC method. It outperforms other baseline models, including two state-of-the-art deep learning classifiers, under the time-varying channel conditions especially when there are mismatched channel conditions of the training set and testing set, which often occur in the real-world. IDK-OGD is also the first online classifier with linear time complexity in AMC, as far as we know.

The insight has a wider implication. It opens up a whole new research area in examining the use of distributional measures, both existing and new, in the AMC task, which has not been considered thus far.

\backmatter

\section*{Declarations}

\bmhead{Conflict of interest}

The authors have no conflicts of interest 
to disclose.

\bmhead{Ethics approval}

Not applicable.

\bmhead{Funding}

Kai Ming Ting is supported by the National Natural Science Foundation of China (Grant number 62076120).

\bmhead{Data availability}

The RadioML2018.01A dataset adopted in this work is open source and available at \url{https://www.deepsig.ai/datasets}.

\bibliography{references}

\end{document}